\pdfoutput=1

\documentclass[11pt]{article}

\usepackage{acl}

\usepackage{times}
\usepackage{latexsym}

\usepackage[T1]{fontenc}

\usepackage[utf8]{inputenc}

\usepackage{amsmath,array,graphicx}
\usepackage{lipsum}
\usepackage{cleveref}
\crefname{section}{§}{§§}
\usepackage{microtype}
\usepackage{tabularx}
\usepackage{multirow}
\usepackage{booktabs}
\usepackage{arydshln}
\definecolor{cadmiumgreen}{rgb}{0.0, 0.42, 0.24}
\usepackage{paralist}
\usepackage[para]{footmisc}
\usepackage{natbib}
\usepackage{xspace}
\usepackage{adjustbox}
\usepackage{tikzit}

\tikzstyle{new style 0}=[fill=white, draw={rgb,255: red,173; green,175; blue,191}, shape=circle, minimum width=.5cm, ultra thick]
\tikzstyle{new style 1}=[fill=white, draw=black, shape=rectangle]
\tikzstyle{new style 2}=[fill=white, draw=red, shape=rectangle]

\tikzstyle{new edge style 0}=[->, fill=none, draw=red, ultra thick]
\tikzstyle{new edge style 1}=[<-, fill=none, ultra thick]
\tikzstyle{new edge style 2}=[draw=black, fill=none, dashed, <-, ultra thick]
\tikzstyle{new edge style 3}=[<-, ultra thick]
\tikzstyle{new edge style 4}=[->, ultra thick]
\tikzstyle{new edge style 5}=[draw=black, fill=none, dashed, ->, ultra thick]

\title{Enhancing Tabular Reasoning with Pattern Exploiting Training}

\author {
    Abhilash Reddy Shankarampeta\textsuperscript{\rm 1\thanks{ Equal Contribution}},
    Vivek Gupta\textsuperscript{\rm {2*}\thanks{ Corresponding Author}},
    Shuo Zhang\textsuperscript{\rm {3}}
    \\
    \textsuperscript{\rm 1}IIT Guwahati;
     \textsuperscript{\rm 2}University of Utah;
     \textsuperscript{\rm 3}Bloomberg
     \\
    sareddy53@gmail.com; vgupta@cs.utah.edu; szhang611@bloomberg.net
}

\newcommand{\datasetName}{{\sc InfoTabS}\xspace}
\newcommand{\alphaOne}{$\alpha_1$\xspace}

\begin{document}
\maketitle
\begin{abstract}

Recent methods based on pre-trained language models have exhibited superior performance over tabular tasks (e.g., tabular NLI), despite showing inherent problems such as not using the right evidence and inconsistent predictions across inputs while reasoning over the tabular data \citep{DBLP:journals/corr/abs-2108-00578}. In this work, we utilize Pattern-Exploiting Training (PET) (i.e., strategic MLM) on pre-trained language models to strengthen these tabular reasoning models' pre-existing knowledge and reasoning abilities. Our upgraded model exhibits a superior understanding of knowledge facts and tabular reasoning compared to current baselines. Additionally, we demonstrate that such models are more effective for underlying downstream tasks of tabular inference on \datasetName. Furthermore, we show our model's robustness against adversarial sets generated through various character and word level perturbations.

\end{abstract}




\section{Introduction} 
\label{introduction}
Natural Language Inference (NLI) is the problem of categorizing a hypothesis into entailment, contradiction, or neutral based on the given premise \cite{DBLP:series/synthesis/2013Dagan}. Large language models such as BERT \cite{devlin-etal-2019-bert}, RoBERTa \cite{DBLP:journals/corr/abs-1907-11692} have been applied to large datasets like SNLI \cite{bowman-etal-2015-large}, MultiNLI \cite{N18-1101}, where they have shown performance comparable to that of humans. 




However, the existing methods based on language models are ineffective for reasoning over semi-structured data \cite{DBLP:journals/corr/abs-2108-00578}. These models often ignore relevant rows and use spurious correlations in hypothesis or pre-training information for making inferences \cite{neeraja-etal-2021-incorporating,poliak-etal-2018-hypothesis,gururangan-etal-2018-annotation,jain-etal-2021-tabpert,DBLP:journals/corr/abs-2108-00578}. Due to existing biases in human curated datasets \cite{rajpurkar-etal-2018-know,zhou-bansal-2020-towards} with hypothesis having annotation artifacts \citep{gururangan-etal-2018-annotation}, often models trained on such data lack generalizability and robustness \citep{glockner-etal-2018-breaking}. Furthermore, the absence of comprehensive test sets hinders robust model evaluation. Thus, evaluating models based only on accuracy does not reflect their reliability and robustness \cite{ribeiro-etal-2020-beyond,moradi-samwald-2021-evaluating}.


\begin{table}
\small
\centering
\begin{tabularx}{\linewidth}{c c}
 \toprule
 \multicolumn{2}{c}{\textbf{Breakfast in America}} \\
 \midrule
 \multicolumn{1}{l}{\textbf{Released}} & \multicolumn{1}{r}{29 March 1979}\\
\multicolumn{1}{l}{\textbf{Recorded}} & \multicolumn{1}{r}{May–December 1978}\\
\multicolumn{1}{l}{\textbf{Studio}} & \multicolumn{1}{r}{The Village Recorder in LA}\\
\multicolumn{1}{l}{\textbf{Genre}} & \multicolumn{1}{r}{Pop, art rock, soft rock}\\
\multicolumn{1}{l}{\textbf{Length}} & \multicolumn{1}{r}{46:06}\\
\multicolumn{1}{l}{\textbf{Label}} & \multicolumn{1}{r}{A\&M}\\
\multicolumn{1}{l}{\textbf{Producer}} & \multicolumn{1}{r}{Peter Henderson, Supertramp}\\
 \bottomrule\\
 \multicolumn{2}{l}{\textcolor{cadmiumgreen}{\textbf{H1}}: Breakfast in America is a pop album with a duration}\\
 \multicolumn{2}{l}{ less than 50 minutes.}\\
 \multicolumn{2}{l}{\textcolor{gray}{\textbf{H2}}: Peter Henderson produces only rock albums.}\\
 \multicolumn{2}{l}{\textcolor{red}{\textbf{H3}}: Breakfast in America was released towards the end}\\
  \multicolumn{2}{l}{of 1979.}\\
  \multicolumn{2}{l}{\textcolor{cadmiumgreen}{\textbf{H4}}: Breakfast in America is recorded in California.}\\
  \multicolumn{2}{l}{\textcolor{gray}{\textbf{H5}}: Supertramp is an English band.}\\
  \multicolumn{2}{l}{\textcolor{red}{\textbf{H6}}: The album was released on 29 March 1978.}\\
\end{tabularx}
\caption{An example of tabular premise from \datasetName~ \cite{gupta-etal-2020-infotabs}. The hypotheses
\textcolor{cadmiumgreen}{\textbf{H1, H4}} is entailed, \textcolor{gray}{\textbf{H2, H5}} is a neutral and \textcolor{red}{\textbf{H3, H6}} is a contradiction. Here, the \textbf{bold} entries, which correspond to the first column, are the keys, while the corresponding entries in the second column of the same row are their respective values.}
\label{tab:example}
\end{table}





In this paper, we investigate the current model's reasoning capability, particularly whether they 
can extract the right knowledge and correctly make rational inferences from that extracted knowledge. We focus on the task of tabular reasoning through table inference on \datasetName~ \cite{gupta-etal-2020-infotabs}. For instance, in \cref{tab:example}, a model must filter out the relevant rows, i.e., extract knowledge, before applying the proper reasoning to categorize H1. Reasoning steps can be complex when involving numerical reasoning like count, sort, compare, arithmetic (H1: 46 < 50), commonsense knowledge (H3: December occurs at the end of the year), and factual knowledge (H4: LA is short for Los Angeles). 








It has been proven that LMs pre-trained without explicit supervision on a huge corpus of free web data implicitly incorporate several types of knowledge into their parameters \citep{peters-etal-2019-knowledge}. For extracting this knowledge from language models (LM), various methods utilize probing \citep[and others]{hewitt-liang-2019-designing,voita-titov-2020-information}, attention \cite{jain-wallace-2019-attention,wiegreffe-pinter-2019-attention}, and prompting \citep[and others]{petroni-etal-2019-language,shin-etal-2020-autoprompt} strategies. This internalized knowledge cannot be retrieved when fine-turning for a subsequent task. One explanation is that the objectives of pre-training and fine-tuning are vastly different. This variation in training objectives also diminishes the expected performance gains of the task, hence necessitating further pre-training on training data \cite{DBLP:conf/iclr/XiongDWS20,roberts-etal-2020-much,eisenschlos-etal-2020-understanding}. Therefore, reframing the subsequent task as a joint pre-training objective becomes essential. Hence, we reformulate the tabular NLI, i.e., our downstream task as a cloze-style problem, a.k.a, a mask language modeling (MLM) problem. For fine-tuning, we utilize the efficient Pattern-Exploiting Training (PET) technique \cite{schick-schutze-2021-exploiting,schick-schutze-2021-just,tam-etal-2021-improving}. PET entails establishing pairs of cloze question patterns and verbalizers that enable subsequent tasks to utilize the knowledge of the pre-trained language models. In addition, PET does not need model upgrades, such as adding more layers or parameters during pre-training.  







Compared to direct fine-tuning-based techniques, i.e., training a classifier layer on top of LM, our method improved +8.1 and +25.8 on factual and relational knowledge evaluation tasks, respectively (see \cref{tab:top1drr4_factrelknow}). On \datasetName~, a tabular inference dataset, our PET training approach outperforms +1.72 on $\alpha_1$ (similar to dev), +2.11 on $\alpha_2$ (adversarial set), and +2.55 on $\alpha_3$ (zero-shot set), see \cref{reasoningmlm_results}) the existing baselines.
This shows the effectiveness of our approach, especially on adversarial and out-of-domain challenging instances. Furthermore, we evaluate our improved model against instance perturbations to examine its robustness. These perturbations are generated by modifying existing \datasetName instances, namely by changing names, numbers, places, phrases (paraphrasing), and characters (spelling errors). In addition, we also incorporated counterfactual instances (i.e., negation) to evaluate the model's robustness against pre-trained knowledge overfitting.  The improvement in the counterfactual setting demonstrates that our approach benefits the model to ground better with premise table evidence.




 Our main contributions are the following:

\begin{itemize}

\item We propose a method for generating prompts for determining if current models can infer from knowledge. 

\item We enhance the model's reasoning via prompt learning, i.e., PET, to extract knowledge from semi-structured tables. 


\item Our experiments on \datasetName show that our proposed approach preserves knowledge and improves performance on downstream NLI tasks. The results are robust when assessed on multiple curated adversarial test sets.


\end{itemize}

\noindent
The dataset and associated scripts, are available at
\url{https://infoadapet.github.io/}.

\section{Motivation}
\label{motivation}


\paragraph{Case for Reasoning on Semi-structured Data.}
Reasoning semi-structured data acquire skills such as arithmetic and commonsense, understanding the text types in the tabular cells, and aggregating information across numerous rows if necessary.
For example, to judge the H1 in \cref{tab:example}, the model needs to understand \textit{"duration"} and  \textit{"length"} are the same in the context of the table, which is about a music album. Also, numerical reasoning is required to compare \textit{"46:06" minutes"} is less than \textit{"50 minutes"}. At the same time, the model should understand that the premise (table) is about a music album, so to classify the H1 model needs to understand the information present in 2 rows (\{\textit{"Genre", "Length"}\}) and perform numerical reasoning on top of that factual information. 

\paragraph{Implicit Knowledge is Required for Reasoning.}
For instance, for H3 in \cref{tab:example}, the model needs to first extract the relevant row, i.e., \textit{"Released"} row from the table, then compares the phrase \textit{"end of 1979"} with the "\textit{Released}" row value \textit{"29 March 1979"} implicitly. 
The model needs to perform temporal reasoning to know that \textit{"year 1979"} is correct. However, the month \textit{"March"} is not the \textit{"end of the year"}, but \textit{"November"} or \textit{"December"} is (implicit commonsense temporal knowledge). 
While previous works tried to incorporate knowledge via pre-training \cite{eisenschlos-etal-2020-understanding,neeraja-etal-2021-incorporating}. In this work, we integrate knowledge and reasoning ability simultaneously using Pattern Exploiting Training \cite{tam-etal-2021-improving}. This approach improves the existing knowledge and enhances reasoning compared to existing methods.


\paragraph{Robustness is Critical for Model Evaluation.} 
Tabular reasoning models typically fail on modest input modification, a.k.a. adversarial manipulation of inputs, highlighting the model's poor robustness and generalizability limit~\citep{DBLP:journals/corr/abs-2108-00578}.
Thus, evaluating reasoning models on adversarial sets generated by minimal input perturbation becomes vital. 
As a result, we propose additional adversarial test sets, such as using character and word level perturbations to evaluate various aspects of model understanding and reasoning over tables. 
For example, if H1 (\cref{tab:example}) is changed to \textit{"Breakfast in Wales is a pop album with a duration of fewer than 50 minutes."} now the label of hypothesis H1 is changes from \textbf{entailment} to \textbf{neutral} since we do not know any information of \textit{"Breakfast in Wales"} from \cref{tab:example}. 
These minor input perturbations can alter the hypothesis' semantic interpretation. Idealistically, a robust model with superior reasoning ability should perform well on these input perturbed adversarial sets, as our technique also demonstrates. 



\section{Our Approach}
\label{approach}

\begin{figure*}[t]
  \includegraphics[width=\textwidth]{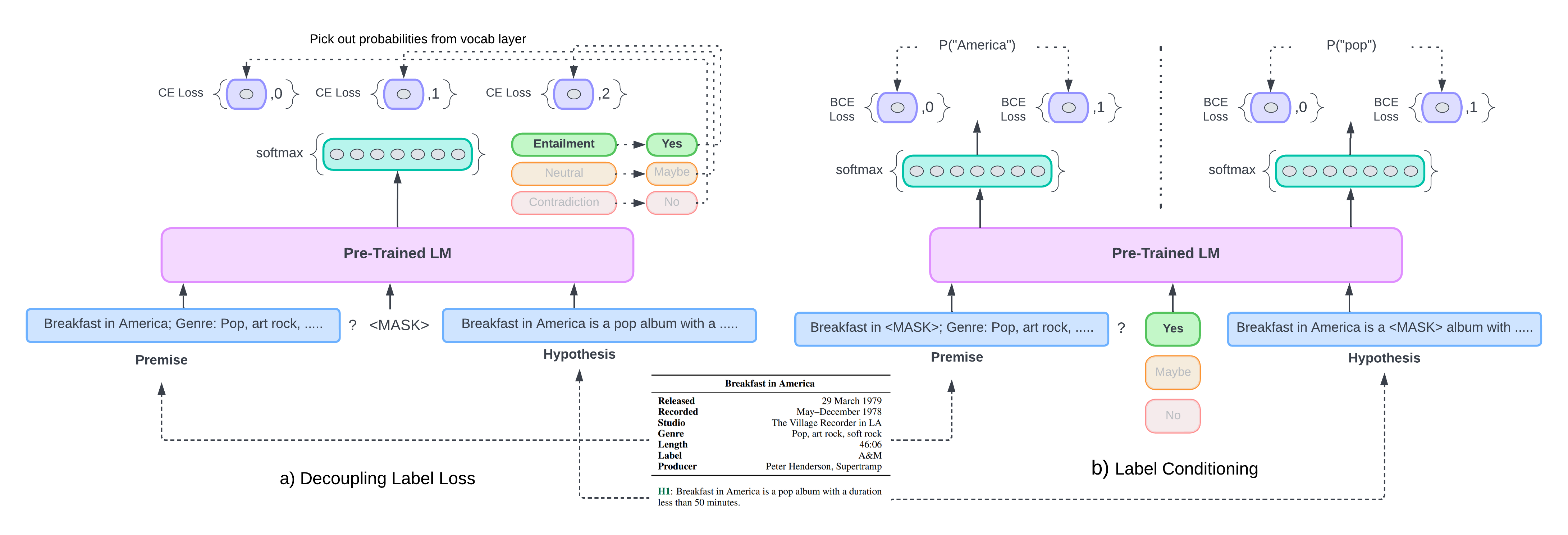}
  \caption{The training uses the two ADAPET components. Here, the blue boxes represent the task inputs (entailed, in this case) a) Decoupling Label Loss: Using the cross entropy loss across all labels, the model must predict the right and wrong labels at the masked-out position. b) Label Conditioning: The model should predict the original token at a randomly masked-out position if the input text has the entail label. Otherwise, not if the label is contradiction or neutral.}
\end{figure*}

In this section we describe our method to \textbf{(a)} evaluate pre-trained LM knowledge for tabular reasoning, \textbf{(b)} enhance model tabular reasoning capability using PET training, \textbf{(c)} and assess model robustness to input perturbations. 

\subsection{Evaluation of Pre-training Knowledge}
To examine how pre-training affects knowledge-based reasoning for tabular data, we focus on two types of knowledge \begin{inparaenum}[(a.)]\item factual knowledge (awareness of specific factual knowledge about entities), \item and relational knowledge (awareness of possible right relations between two distinct entities). \end{inparaenum} For instance, in the sentence \textit{"Breakfast in America was released on March 29, 1979"}, \textit{"Breakfast in America"} and  \textit{"March 29, 1979"} are considered as factual knowledge, while their relationship term, i.e., \textit{"released"} corresponds to relational knowledge.

We evaluate factual and relational knowledge in the language model before and after training for the downstream task like reasoning. In specific, we query the model using "fill-in-the-blank" cloze statements (a.k.a. prompts). As gauging knowledge using prompts is limited by how the prompts are constructed. 
We use part-of-speech tagging to detect nouns and verbs that are then used to mask names, numbers, and dates. These prompts are generated using hypotheses from the $\alpha_1$, and dev sets as these sets have similar distribution as the training data \cite{gupta-etal-2020-infotabs}. We construct the prompts from both entailed and contradictory hypotheses. For prompts derived from entailed hypotheses, the model must predict the correct masked word, i.e., a term semantically equivalent to the word in the hypothesis. In contrast, for the prompts derived from contradicting hypotheses, the model should predict a semantically different term with the same entity type as the one mentioned in the hypothesis. To study the effect of the premise, we also query the model with the premise. To do this we modify the input as \textit{premise + prompt}. 


\paragraph{Prompts for Factual Knowledge Evaluation}

As most factual knowledge is contained in proper nouns and numbers, we randomly mask proper nouns or numbers in the hypothesis to generate a prompt and query the Language Model to fill the masked tokens. For example \textit{"Duration of Breakfast in America is 46 minutes"} (\cref{tab:example}), \textit{"Breakfast in America"}, \textit{46} are the factual information present in the sentence and they are connected by \textit{"duration"}. We randomly mask either \textit{"Breakfast in America"} or \textit{"46"} to generate prompt \textit{"Duration of Breakfast in America is <mask> minutes"}. Occasionally, a masked term can be a number in numeric form (e.g., 2); however, the model predicted word form ("two"). We solved this issue by converting the predicted word into its numeric form or vice versa. E.g. \textit{"Breakfast in America is produced by <mask> producers"}, where \textit{<mask> = two}.



\paragraph{Prompts for Relational Knowledge Evaluation.}
Similar prompts are leveraged for relational knowledge.
For example, to predict \textit{<mask> = released} for \textit{"Breakfast in America was <mask> towards the end of 1979"}, the model needs to understand that \textit{"Breakfast in America"} is a music album to predict \textit{"released"} instead of \textit{"eaten"} which is highly probable due the neighbor context term \textit{"Breakfast"}.
We also use WordNet \cite{10.1145/219717.219748} to discover synonyms for the masked term and see if the predicted word is among them.




\subsection{Knowledge Incorporation for Reasoning}
The issue of deducing inferences from tabular premises is similar to the typical NLI problem, except that the premises are tables rather than sentences. When evaluating the reasoning skills, we use a variety of representations of the tabular premise (see section \ref{tabrep}, \cref{tab_representation}). We also study the effect of pretraining on an NLI task on \datasetName. 


\paragraph{Pattern-Exploiting Training.}
Using Pattern-Exploiting Training (PET) \cite{schick-schutze-2021-exploiting}, NLU tasks are reformulated as cloze-style questions, and fine-tuning is performed using gradient-based methods. We use ADAPET (A Densely-supervised Approach to Pattern-Exploiting Training) \cite{tam-etal-2021-improving}, which increases supervision by separating the label token losses and applying a label-conditioned masked language modeling (MLM) to the entire input. 

The input to the language model is converted into a cloze-style form with the pattern  \textit{<premise> ? <mask>, <hypothesis>}. The model is tasked to predict the masked word from the vocabulary. The model computes each token's probability as a softmax normalized overall tokens, allowing the logits of all vocabulary tokens to impact each likelihood, similar to the regular MLM objective. While in PET, the masked word is forced to predict from the output space \textit{\{Yes, Maybe, No\}} which are mapped to labels \textit{\{Entailment, Neutral, Contradiction\}}. As a result, there will never be a gradient signal for non-label tokens. Inverting the query to the model to \textit{"In light of the answer, what is the appropriate context?"} from \textit{"What is the appropriate label based on the input?"} label conditioned mask language modeling is introduced by randomly masking out context tokens. If the label is "entail", during training, the model is obligated to predict the original token; however, if the label is "contradiction" or "neutral", the model is forced to ignore the original token.

\paragraph{Masked Language Modeling.}
ADAPET randomly masks tokens (RoBERTa style) from the context. Inspired by SpanBERT \cite{joshi-etal-2020-spanbert}, ERNIE \cite{DBLP:journals/corr/abs-1904-09223}, we sample and mask the entire words based on pre-defined conditions. In Conditional Whole Word Masking (CWWM), we create a set of words $S_{w}$ from a given sentence, and the POS of the words in that set must be from \{"Adjective", "Adverb", "Noun, "Verb", "Proper Noun", "Adposition", "Numeral", "Coordinating Conjunction", "Subordinating Conjunction" \}\footnote{\url{https://universaldependencies.org/u/pos/}}. We sample words from the set $S_{w}$ and mask all tokens matching the sampled word concurrently while maintaining the same overall masking rate.


%

\begin{table*}
\small
\centering
\begin{tabularx} {\linewidth}{l | l | l}
\toprule
\multicolumn{1}{l}{Perturbation} & \multicolumn{1}{c}{Original text} & \multicolumn{1}{c}{Perturbed text} \\ \midrule
\multirow{4}{*}{\textbf{Character}} & \multirow{4}{*}{Peter Henderson produces only rock albums} & \multicolumn{1}{l}{Peter Hen\textcolor{blue}{bg}derson produces only rock alb\textcolor{blue}{s}ums} \\

 & & \multicolumn{1}{l}{Peter He\textcolor{blue}{nd}ers\textcolor{blue}{no} produces only ro\textcolor{blue}{kc} albums} \\
 
 & & \multicolumn{1}{l}{\underline{Pter} Henderson produces \underline{onl} rock \underline{abus}} \\
 
 & & \multicolumn{1}{l}{Pet\textcolor{blue}{q}r Hen\textcolor{blue}{k}erson pr\textcolor{blue}{g}duces only rock al\textcolor{blue}{oc}ms} \\ 
 \hdashline
 \multirow{3}{*}{\textbf{Location}} & \multicolumn{1}{l|}{Breakfast in America is recorded in California} & \multicolumn{1}{l}{Breakfast in America is recorded in \textcolor{blue}{Florida}.} \\
  & \multicolumn{1}{l|}{Breakfast in America is recorded in USA} & \multicolumn{1}{l}{Breakfast in America is recorded in \textcolor{blue}{Syria}.} \\
  & \multicolumn{1}{l|}{Breakfast in America is by an English rock band.} & \multicolumn{1}{l}{Breakfast in America is by an \textcolor{blue}{Mexican} rock band.} \\
 \hdashline
 \multirow{1}{*}{\textbf{Name}} & \multicolumn{1}{l|}{Peter Henderson produces only rock albums} & \multicolumn{1}{l}{\textcolor{blue}{John Doe} produces only rock albums} \\
  \hdashline
 \multirow{2}{*}{\textbf{Numbers}} &  \multirow{2}{*}{The album was released on 29 March 1978.} & \multicolumn{1}{l}{The album was released on 29 March \textcolor{blue}{346}.} \\
 & & \multicolumn{1}{l}{The album was released on \textcolor{blue}{1} March 1978.} \\
 \hdashline
 \multirow{1}{*}{\textbf{Negation}} & \multicolumn{1}{l|}{The genres of the album are pop and rock.} & \multicolumn{1}{l}{The genres of the album are \textcolor{blue}{not} pop and rock.} \\
  \hdashline
 \multirow{1}{*}{\textbf{Paraphrase}} & \multicolumn{1}{l|}{The album was recorded in the last half of 1979.} & \multicolumn{1}{l}{\textcolor{blue}{In the second part of 1979, the album was recorded.}} \\

 \bottomrule
 \end{tabularx}
\caption{Examples of various perturbations used to generate the adversarial test sets based on \cref{tab:example}.} 
\label{tab:data_adv_examples}
\end{table*}

\subsection{Robustness with Input Perturbations} 




We apply a range of character- and word-level perturbations to hypotheses to simulate circumstances where the input is slightly noisy or deviates from the training data distribution. We use TextAttack \cite{morris-etal-2020-textattack}, NLP Checklist \cite{ribeiro-etal-2020-beyond}, and manual perturbations for generating the adversarial data. These adversarial sets will test the dependence of the model on word overlap, numerical comprehension, and hypothetical assertions. Refer to  \cref{tab:data_adv_examples,tab:data_adv_examples_more} for examples.  

\noindent
\textbf{Character-level perturbation} 
employs perturbations such as introducing random characters, switching characters, removing a random character, and substituting a random character in the randomly selected word.
This alteration does not impact the label of the hypothesis because it does not alter the sentence's meaning.

\noindent
\textbf{Location perturbation} modifies the identified locations (countries, cities, and nationalities) in a sentence to another place specified in the location map. 
The NER model (TextAttack) identifies the location in a given sentence and replaces it with a sampled location from a dictionary. Here, cities are replaced with other cities and similar changes for countries.
This perturbation transforms the entail clauses into contradictions but does not affect the original neutral and contradiction labels.



\noindent
\textbf{Name perturbation} randomly replaces a person's name with the other one from a name list.
This perturbation alters the label of every hypothesis into a neutral because the perturbed hypothesis and premise mention different persons.

\begin{table}[ht]
\small
\centering
\begin{tabular}{l r |l r} 
 \toprule
  \textbf{Peturb Type} &\textbf{Size} & \textbf{Peturb Type} &\textbf{Size}\\ \midrule
character & 1800 & negation+char & 1726 \\
location & 1229 & negation+name & 1677 \\
name & 1646 & number+char & 837 \\
negation & 1726 & number+name & 776 \\
number & 837 & number+negation & 817\\
paraphrase & 1800 & num+paraphrase & 837\\
num+para+name & 776 & paraphrase+name & 1721 \\
 \bottomrule
\end{tabular}
\caption{Number of examples for each perturbation type in the adversarial set.}
\label{tab:advset_size}
\end{table}

\noindent
\textbf{Perturbing Numbers} changes the entailed sentences into contradictions but does not affect the labels of neutral and contradictions.
Contradictory statements remain contradictory because it is implausible that a randomly sampled number will be the actual number in the premise, making the hypothesis entailed.


\noindent
\textbf{Negation} transforms entailment into a contradiction by negating the given sentence, keeping neutrals intact.

\noindent
\textbf{Paraphrasing}
paraphrases the given sentences without the loss of meaning using manual paraphrasing and Pegasus model\footnote{\url{https://biturl.top/MzQnMv}}. Paraphrasing does not affect the inference label as it does not change the semantic meaning of the hypothesis.


\noindent
\textbf{Composition of Perturbations} perturbs sentences by applying various distinct perturbations sequentially. E.g., in \textbf{num+para+name} we perturbed a sentence \textit{"Supertramp, produced an album that was less than 60 minutes long"}, with premise \cref{tab:example} to \textit{"Supertramp, produced an album that was less than \textcolor{blue}{40} minutes long"} (number) then \textit{"Supertramp \textcolor{blue}{released} an album \textcolor{blue}{which lasted} less than 40 minutes."} (paraphrase) then \textit{\textcolor{blue}{"James} released an album which lasted less than 40 minutes"} (name). 
\section{Experiments and Analysis}
\label{experiments}



\textbf{Dataset.} \label{para:dataset}Our experiments we use \datasetName, a tabular inference dataset introduced by \citet{gupta-etal-2020-infotabs}. The dataset is diverse in terms of the tables domains, categories, and corresponding keys (entity types and forms) it contains, as illustrated in examples table \ref{tab:example}. In addition, \citet{gupta-etal-2020-infotabs} reveals that inference on corresponding hypotheses requires extensive knowledge and commonsense reasoning ability. 
Given the premise table, hypothesis in the dataset is labeled as either an Entailment (\textcolor{cadmiumgreen}{E}), Contradiction (\textcolor{red}{C}), or Neutral (\textcolor{gray}{N}). 

In addition to the conventional development set and test set (referred to as $\alpha_1$), an adversarial test set ($\alpha_2$) lexically equivalent to $\alpha_1$ but with minor changes in the hypotheses to flip the entail-contradict label and a zero-shot cross-domain test set ($\alpha_3$) containing large tables from other domains that are not in the training set are used for evaluation. For all of our experiments, we use the accuracy of classifying the labels as our primary metric for evaluation. The domain of tables in training sets and $\alpha_1$,$\alpha_2$ are similar. However, the training and fine-tuning tables are exclusive. Each of the test sets $\alpha_1,\alpha_2,\alpha_3$ has 200 unique tables paired with 9 hypothesis sentences (3\textcolor{cadmiumgreen}{E}, 3\textcolor{red}{C}, 3\textcolor{gray}{N}), totalling 1800 table-hypothesis pairs. Table \ref{tab:advset_size} depict the statistics of perturbed sets from \datasetName.

\paragraph{Model.} We use the pre-trained RoBERTa-Large (RoBERTa$_L$) \cite{DBLP:journals/corr/abs-1907-11692} language model from HuggingFace \cite{wolf-etal-2020-transformers} for all of our investigations. We employ various configurations of language models to assess knowledge in two different cases. These configurations include RoBERTa$_L$, RoBERTa$_L$ finetuned on \datasetName~ (RoBERTa$_L$+CLS), RoBERTa$_L$ trained for tabular inference using PET (ADAPET), and finetuning \datasetName~ on ADAPET (ADAPET+CLS). Here we define finetuning as training a classifier head (CLS). We also investigate the effect of NLI pre-training using RoBERTa$_L$ pretrained on MNLI \cite{N18-1101}, and mixed dataset (mixNLI) containing ANLI+MNLI+SNLI+FeverNLI  \footnote{\url{https://biturl.top/e6Vney} \label{mixnli}} \cite{nie-etal-2020-adversarial,bowman-etal-2015-large,10.1609/aaai.v33i01.33016859}. All models are trained on 16538 table-hypothesis pairs (1740 tables) for 10 epochs with a 1e-5 learning rate.


\paragraph{Table Representation.} \label{tabrep} We explored two ways to represent table  (a.) \emph{Table as paragraph} uses Better Paragraph Representation for table representation, (b.) and \textit{Distracting Row Removal} prunes tables based on the similarity between hypothesis and tables rows. We investigated the pruning of top 4 (DRR@4) and top 8 (DRR@4) rows for our experiments. Both representation methods are adapted from \citet{neeraja-etal-2021-incorporating}. For more details on table representation, refer to \cref{tab_representation}. 

\subsection{Results and Analysis} 
\label{sec:results}

Our experiments answer the following questions:

\paragraph{RQ1:} Can the large language model use pre-trained knowledge for reasoning? Does our adaptive training method enhance model reasoning?

\paragraph{RQ2:} Does fine-tuning downstream tasks benefit model reasoning? Can our adaptive training benefit model via enhancing its reasoning knowledge?

\paragraph{RQ3:} Is our adaptive method-based model robust to input perturbations? Can our method enhance model's semantic-syntactic comprehension?



\paragraph{Models Knowledge Evaluation.}
To answer RQ1, we evaluate the knowledge in the presence and absence of the premise using the Entail and Contradictory hypotheses, which are taken from the evidence in the premise tables. We do not use Neural statements as they may contain subjective and out-of-table information.

\begin{table}[!htbp]
\small
\centering
\setlength{\tabcolsep}{4.0pt}
\begin{tabularx}{\columnwidth}{l @{\hspace{1.2\tabcolsep}} l c r r r} 
 \toprule
 \textbf{Type} &\textbf{Input} &\multicolumn{2}{c}{\textbf{RoBERTa$_L$}} &\multicolumn{2}{c}{\textbf{ADAPET}} \\ \midrule
 \multicolumn{2}{c}{\bf Top 1 Accuracy} &\textbf{w/o} &\textbf{+CLS} &\textbf{w/o} &\textbf{+CLS}
\\\midrule
\multirow{6}{*}{Factual} &only \textcolor{cadmiumgreen}{E} &35.5 &26.2 &34.3 &29.2 \\
&prem + \textcolor{cadmiumgreen}{E} &59.4 &29 &59.7 &44.8 \\
&only \textcolor{red}{C} &37.2 &24.6 &36.9 &29.8 \\
&prem + \textcolor{red}{C} &54.6 &26.5 &49.7 &39.9 \\
&only \textcolor{cadmiumgreen}{E}$\cup$\textcolor{red}{C} &36.3 &25.4 &35.5 &29.5 \\
 &prem + \textcolor{cadmiumgreen}{E}$\cup$\textcolor{red}{C} &57.7 &27.8 &54.6 &42.5 \\ \hdashline
\multirow{6}{*}{Relational} &only \textcolor{cadmiumgreen}{E} &48.9 &27 &52.8 &35.6 \\
&prem + \textcolor{cadmiumgreen}{E} &57.7 &22.4 &58.7 &41 \\
&only \textcolor{red}{C} &44.7 &27.3 &47.3 &35.6 \\
&prem + \textcolor{red}{C} &51.8 &24 &52.9 &34 \\
&only \textcolor{cadmiumgreen}{E}$\cup$\textcolor{red}{C} &46.7 &27.2 &49.9 &35.6 \\
&prem + \textcolor{cadmiumgreen}{E}$\cup$\textcolor{red}{C} &54.6 &23.2 &55.7 &37.3
\\

 \bottomrule
\end{tabularx}
\caption{Top 1 accuracy of Factual \& Relational Knowledge Evaluation on DRR@4.(w/o - no CLS, RoBERTa$_L$+CLS}
\label{tab:top1drr4_factrelknow}
\end{table}


\begin{table*}[!htbp]
\centering
\setlength{\tabcolsep}{4.0pt}
\footnotesize
\begin{tabular}{lcccccccccc}
\toprule
\textbf{Splits} &\textbf{Premise} &\textbf{RoBERTa$_L$} &\multicolumn{4}{c}{\textbf{ADAPET}} &\multicolumn{4}{c}{\textbf{ADAPET+CLS}} \\
\cmidrule(lr){4-7}\cmidrule(lr){8-11}
 & &\textbf{+CLS} &\textbf{token} &\textbf{CWWM} &\textbf{+mixNLI} &\textbf{+MNLI} &\textbf{token} &\textbf{CWWM} &\textbf{+mixNLI} &\textbf{+MNLI} \\\midrule
\multirow{3}{*}{Dev} &BPR &76.83 &77.5 &77.67 &79.07 &78.07 &77.66 &77.27 &\textbf{79.63} &78.46 \\
&DRR@4 &76.39 &76.67 &76.97 &78.57 &77.33 &76.88 &77.11 &\textbf{78.64} &77.44 \\
&DRR@8 &75.36 &77.77 &77.63 &78.83 &77.93 &77.81 &77.57 &\textbf{79.42} &78.96 \\\midrule
\multirow{3}{*}{$\alpha_1$} &BPR &75.29 &76.87 &75.93 &77.33 &77.47 &77.47 &78.05 &77.96 &\textbf{78.33} \\
&DRR@4 &75.78 &77.5 &77.53 &\textbf{78.6} &78.17 &77.18 &77.66 &78.04 &78.13 \\
&DRR@8 &75.61 &78.3 &78 &79 &78.2 &78.03 &78.7 &78.63 &\textbf{79.05} \\\midrule
\multirow{3}{*}{$\alpha_2$} &BPR &66.5 &67.93 &68.07 &\textbf{72.4} &69.8 &68.48 &69.55 &72.16 &70.09 \\
&DRR@4 &67.22 &69.33 &69 &70.23 &69.03 &68.92 &68.29 &\textbf{70.58} &69.24 \\
&DRR@8 &67.11 &69.43 &69.37 &71.87 &69.97 &69.24 &69.81 &\textbf{72.13} &70.61 \\\midrule
\multirow{3}{*}{$\alpha_3$} &BPR &64.26 &63.73 &64.6 &66.23 &64.13 &64.98 &65.67 &\textbf{68.4} &66.03 \\
&DRR@4 &64.88 &67.43 &67.5 &68.7 &67.33 &66.02 &66 &\textbf{68.74} &67.37 \\
&DRR@8 &67.53 &68.07 &67.63 &\textbf{70.2} &68 &66.66 &67.59 &69.2 &68.31 \\

\bottomrule
\end{tabular}
\caption{\label{reasoningmlm_results} Reasoning results on \datasetName~ comparing RoBERTa$_L$+CLS, ADAPET, ADAPET+CLS (without pre-training (token, CWWM), with mixNLI, MNLI pre-training). token, CWWM - masking strategies, mixNLI, MNLI pre-training uses RoBERTa style token masking.}
\end{table*}

In all the settings (\cref{tab:top1drr4_factrelknow,tab:top5drr4_factrelknow}) with and without premise, our model outperformed RoBERTa$_L$+CLS. The addition of the premise enhances model performance further. This can be ascribed to additional knowledge in the premise that our PET-trained model can leverage efficiently for reasoning. From \cref{tab:top1drr4_factrelknow}, we observe that for all settings, our approach gave \~100$\%$ improvement in relational knowledge evaluation compared to RoBERTa$_L$+CLS. Even training a classifier on top of ADAPET outperforms RoBERTa$_L$+CLS. We also evaluated on contradiction hypothesis to assess if the model can rightly identify false claims despite having correct entity types. 

There is a significant difference between the Top 1 accuracy of premise+\textcolor{cadmiumgreen}{E} and premise+\textcolor{red}{C} for factual knowledge evaluation as the model should not predict the masked token in the prompt from a contradiction statement, especially in factual prompts. And for relational knowledge, irrespective of the label of the hypothesis, the model should predict the masked token correctly if the model rightly understands the entity types of words in the sentence. In almost all the settings, our approach performs almost comparable to RoBERTa$_L$, and it even outperforms RoBERTa$_L$ in only Entail, and Premise+ Entail settings. Training a classifier on top of RoBERTa$_L$ decreases the performance knowledge evaluation but training a classifier head on top of ADAPET still tops RoBERTa$_L$+CLS, thus demonstrating the benefits of our approach. A similar observation was reported with Top 5 accuracy (\cref{tab:top5drr4_factrelknow}).



\paragraph{Knowledge Incorporation for Reasoning.}

To answer RQ2, we experiment with various premise representations of tables as paragraphs (BPR, DRR@4, DRR@8) (see \cref{reasoningmlm_results}). We observe that Roberta-Large with ADAPET improves performance in all premise representations except for $\alpha_3$ with BPR compared to RoBERTa$_L$+CLS due to an increased number of keys in the tables (13.1 per table in $\alpha_3$ when compared to 8.8 per table in $\alpha_1$ and $\alpha_2$). Results in \cref{reasoningmlm_results} are the average accuracy of the models tested on multiple seeds.

With ADAPET, we also improve performance using linearized table (see \cref{tab:linear_table}) compared to \citet{gupta-etal-2020-infotabs} (+1.04 in $\alpha_1$, +0.58 in $\alpha_2$, +0.69 in $\alpha_3$). ADAPET (token masking, no pre-training) tops RoBERTa$_L$+CLS in every premise representation and test split. +1.72 in $\alpha_1$, +2.11 in $\alpha_2$, +2.55 in $\alpha_3$ with DRR@4. CWWM with ADAPET also outperformed RoBERTa$_L$+CLS. However, the performance of the two masking procedures is comparable for all test sets, even with the classifier setting.


\begin{table*}[!htbp]
\small
\centering
\setlength{\tabcolsep}{4.0pt}
\begin{tabular}{lccccccccc}
\toprule
\textbf{Perturb} &\textbf{RoBERTa$_L$} &\multicolumn{4}{c}{\textbf{ADAPET}} &\multicolumn{4}{c}{\textbf{ADAPET+CLS}} \\
\cmidrule(lr){3-6}\cmidrule(lr){7-10}
 &\textbf{+CLS} &\textbf{token} &\textbf{CWWM} &\textbf{+mixNLI} &\textbf{+MNLI} &\textbf{token} &\textbf{CWWM} &\textbf{+mixNLI} &\textbf{+MNLI}\\\midrule
num+para+name &13.04 &10.1 &7.1 &11.7 &10.1 &11.7 &13.81 &\textbf{16.62} &13.55 \\
number+name &15.72 &14.6 &9.0 &14 &13.2 &15.6 &15.36 &\textbf{18.94} &15.85 \\
negation+name &19.08 &16.1 &7.2 &\textbf{20} &11.6 &14.43 &12.88 &14.37 &12.1 \\
num+paraphrase &27.46 &59.5 &\textbf{61.0} &58.4 &57.3 &52.5 &51.49 &56.63 &54.95 \\
paraphrase+name &30.79 &22.6 &18.3 &28.3 &24.9 &27.01 &27.3 &\textbf{30.85} &27.71 \\
name &32.7 &24.7 &19.0 &31.1 &28 &28.9 &29.96 &\textbf{33.44} &30.69 \\\hdashline
random &33.33 &33.33 &33.33 &33.33 &33.33 &33.33 &33.33 &33.33 &33.33\\ \hdashline
number+negation &36.13 &42.7 &31.8 &\textbf{53.2} &28.3 &37.91 &47.32 &37.75 &24.04 \\
negation+char &39.39 &41.4 &38.5 &\textbf{47.6} &40.1 &42.9 &41.94 &42.06 &40.85 \\
negation &53.7 &58.1 &53.3 &\textbf{64.8} &56.1 &57.6 &56.83 &59.15 &53.88 \\
number+char &54.43 &58.8 &\textbf{65.2} &57.1 &60.3 &55.79 &47.9 &57.1 &59.28 \\
number &56.1 &57.8 &\textbf{62.0} &57.8 &57 &52.44 &51.37 &55.79 &54.6 \\
character &63.05 &62.8 &63.3 &65.9 &64.4 &64.05 &64.44 &66.05 &\textbf{66.83} \\
location &67.6 &70 &\textbf{70.2} &67.7 &69.1 &69.81 &66.8 &67.4 &65.98 \\
paraphrase &70.56 &72.3 &73.2 &\textbf{73.8} &73.4 &71.6 &70.5 &72.66 &72.3 \\\hdashline
\datasetName (\alphaOne) &76.56 &78.1 &78.9 &\textbf{80.2} &78.9 &78.27 &77.66 &78.5 &78.66 \\
\bottomrule
\end{tabular}
\caption{\label{addrr8v_results} Adversarial Reasoning results on perturbed sets with DRR@8 comparing RoBERTa$_L$+CLS, ADAPET, ADAPET+CLS (without pre-training (token, CWWM), with mixNLI, MNLI pre-training), token, CWWM - masking strategies, mixNLI, MNLI pre-training uses RoBERTa style token masking. Rows in the tables are sorted in ascending order w.r.t RoBERTa$_L$+CLS performance.}
\end{table*}

We notice that the DRR@8 representation outperforms the best, especially in $\alpha_3$ due to removing the irrelevant rows (+4.34 over BPR, +0.64 over DRR@4). The zero-shot test set $\alpha_3$ which has a significant proportion of unseen keys (different domain tables) when compared to other test sets (number of unique keys intersection with train is 312, 273, 94 for $\alpha_1$, $\alpha_2$ and $\alpha_3$ respectively) has seen a substantial improvement with the use of NLI pre-trained model. When compared to ADAPET (token masking, no pretraining), there has been an improvement of +2.13 units (no CLS) and +2.54 units (with CLS) with DRR@8 over no pre-training. We also observed that pre-training in more diverse data helps improve performance \citep{andreas-2020-good,pruksachatkun-etal-2020-intermediate}. Models which are pre-trained on mixNLI\footref{mixnli} outperformed MNLI pre-trained in almost every setting (+0.8 in $\alpha_1$, +1.9 in $\alpha_2$, +2.2 in $\alpha_3$ with no CLS, DRR@8).



\paragraph{Robustness to Input Perturbation.}
To answer RQ3, we evaluate our model on several challenging input perturbations. The perturb test sets are generated using various character-level, and word-level perturbations are also tested with BPR, DRR@4, and DRR@8 table representations (see \cref{addrr8v_results}). To generate these sets, we applied perturbations on $dev$, and $\alpha_1$ sets as the distribution of these sets are similar to the training set. We also human-verified our perturbation examples; refer to \cref{appendix_qualit_pertub}.

Except for the perturbations involving names, our method ADAPET (no pre-training) outperforms RoBERTa$_L$+CLS. We see the max improvement of ADAPET in the Negation (+4.4); this implies our model can handle counterfactual statements well. We observed that training a classifier head on top of ADAPET performed better with the adversarial sets involving multiple perturbations. In the challenge set with \textit{number+paraphrase} all the ADAPET-based models outperformed RoBERTa$_L$+CLS by 2x times. We observed that using NLI pre-training also helps substantially improve the robustness. With the use of mixNLI and MNLI pre-trained weights, the performance of ADAPET-based models improved substantially compared to those without pre-training, even outperforming RoBERTa$_L$+CLS. From \cref{addrr8v_results}, it is clear that with hypotheses involving multiple perturbations, RoBERTa$_L$+CLS tends to perform more poorly compared to the ADAPET-based model. (For quality analysis of perturbations see \cref{appendix_qualit_pertub}). The performance on all perturb sets is much worse than that of the corresponding model on dev, $\alpha_1$ sets. Improving the performance of these sets is crucial.

 

\paragraph{What did we learn?} 
Reformulating the NLI task as an MLM problem enabled the inclusion of premise table knowledge into Language Models (LM) for efficient reasoning.
Using ADAPET, we have shown that knowledge can be retained and assimilated into reasoning tasks more effectively. ADAPET training also improves the model's ability to reason on downstream tasks.
Similar observation is also observed in prior works \citet{DBLP:conf/iclr/XiongDWS20,DBLP:journals/corr/abs-1904-09223} where MLM is utilized to incorporate external knowledge, although the later require additional table based pre-training. 
Moreover, \citet{DBLP:journals/corr/abs-2108-00578,lewis-etal-2021-question} have shown that the LM utilizes spurious patterns to accomplish reasoning tasks. 
Our perturb sets study informed us that our ADAPET-based method is more robust than direct classification to semantic-syntactic alternations. (see \cref{discussion} for further discussions)

\section{Related Work}
\label{relwork}

\paragraph{Tabular Reasoning.} Many recent papers discussed NLP challenges associated  with semi-structured table data such as Tabular NLI \cite{gupta-etal-2022-right,gupta-etal-2020-infotabs,neeraja-etal-2021-incorporating}, fact verification \cite{DBLP:conf/iclr/ChenWCZWLZW20,zhang-etal-2020-table}, question answering \citep[ and others]{zhu2021tat,zhang2020tablesurvey,pasupat:15,krishnamurthy2017neural,7845035,10.1145/2872427.2883080,chen-etal-2020-hybridqa,DBLP:journals/corr/abs-2012-14610,lin2020bridging,zayats2021representations,chen2021kace}, and text generation from tables \citep[ and others]{parikh2020totto,zhang2020tablesum, nan-etal-2021-dart,DBLP:conf/iclr/ChenCSWC21,DBLP:journals/corr/abs-2107-07261} are some examples. Several studies have offered techniques for encoding Wikipedia tables, such as TAPAS\citep{herzig-etal-2020-tapas}, TaBERT \citep{yin-etal-2020-tabert}, TabStruc \citep{zhang-etal-2020-table}, TABBIE \citep{iida-etal-2021-tabbie}, StruBERT \citep{trabelsi2022structbert}, Table2Vec \citep{zhang2020table2vec}, TabGCN \citep{pramanick2021joint} and RCI \citep{glass-etal-2021-capturing}, amongst others. Works suchs as \citep[ and others]{yu2018spider,DBLP:conf/iclr/0009WLWTYRSX21,eisenschlos-etal-2020-understanding,neeraja-etal-2021-incorporating, muller-etal-2021-tapas} investigate tabular data augmentation.


\paragraph{Knowledge Incorporation and Evaluation.} 
A line of works have been proposed to integrate knowledge into the LMs using pretrained entity embeddings \citep[and others]{zhang-etal-2019-ernie,peters-etal-2019-knowledge}, external memory \cite{logan-etal-2019-baracks,DBLP:conf/iclr/KhandelwalLJZL20,DBLP:journals/corr/abs-2109-04223}, unstructured text \cite{DBLP:conf/iclr/XiongDWS20,DBLP:journals/corr/abs-1904-09223}. Several methods, including probing classifiers, have been proposed to extract and assess knowledge from LMs \citep[and others]{hewitt-liang-2019-designing,voita-titov-2020-information,DBLP:journals/corr/abs-2202-00964}, attention visualization \cite{jain-wallace-2019-attention,wiegreffe-pinter-2019-attention}, and prompting \cite{petroni-etal-2019-language,shin-etal-2020-autoprompt,jiang-etal-2020-know}. Many works have been published to study and create the prompts \citep[and others]{shin-etal-2020-autoprompt,DBLP:journals/corr/abs-2107-13586,10.1145/219717.219748,qin-eisner-2021-learning}.


\paragraph{Model Robustness.}
Many works proposed ways to evaluate robustness to noise, fairness, consistency, explanation, error analysis, and adversarial perturbations to test the model's robustness and reliability \citep[e.g.,][]{ribeiro2016should, ribeiro2018anchors, ribeiro2018semantically, zhao2018generating, iyyer2018adversarial,glockner-etal-2018-breaking, naik2018stress, mccoy2019right, nie2019analyzing, liu2019inoculation}.
\citet{moradi-samwald-2021-evaluating} introduces a textual perturbation infrastructure that incorporates character- and word-level systematic perturbations to imitate real-world noise. \citet{goel-etal-2021-robustness} offered a toolbox to evaluate NLP systems on subpopulations, transformations, evaluation sets, and adversarial attacks.
\section{Conclusion}
\label{conclusion}

In this work, we have validated the effects of factual and relational knowledge in the language model via handcrafted prompts for tabular reasoning.
Through prompt learning, i.e., Pattern-Exploiting Training, we extracted knowledge from semi-structured tables and further improved the model's reasoning capabilities. Our intensive experiments on the \datasetName demonstrate that our approach can conserve knowledge and enhance tabular NLI performance. The conclusions hold up well when tested against carefully crafted adversarial test sets based on character and word-level perturbations. 

\paragraph{Method Limitations:}
Entity tables are the focus of our solution. 
Its scalability in constructing prompts and other tables with different structures is limited by the idea that manually identified pattern from the specific dataset and template-based prompts. In addition, as not different from other NLP tasks, automatically detecting knowledge patterns and bridging patterns to prompts, especially for semi-structured tables, is under-explored. Furthermore, investigating prompting for sophisticated structured tables such as nested structures (e.g., lists inside tables), hierarchical tables (e.g., table inside a table), and multi-modal tables (pictures within table) will necessitate substantial effort. 

\paragraph{Future Directions:}
We have identified the following future directions:
(a.) \textit{Designing better prompts for knowledge evaluation}: Our current prompts treat entail and contradictory statements as the same while evaluating knowledge. In the presence of the premise, masking \textit{Breakfast in America} in H3 (\cref{tab:example}) and using that as an input model will predict Breakfast in America even though the hypothesis is a contradiction. 
We want to work on developing prompts label conditioned evaluation based on existing work on prompt engineering. \cite{DBLP:journals/corr/abs-2107-13586}.
(b.) \textit{Improving Robustness:} \label{robust_discuss} While our models' performance on the challenging adversarial test sets is lower than benchmarks on \datasetName~, we do not know its reason. The created test sets may be challenging because they focus on phenomena that existing models cannot capture or exploit blind spots in a model's training set. Following the ideas of Inoculation by Fine-Tuning \cite{liu-etal-2019-inoculation}, we want to improve and assess the reasons behind the results in \cref{addrr8v_results}.

\section*{Acknowledgement}
We thank members of the Utah NLP group for their valuable insights and suggestions at various project stages and reviewers for their helpful comments. Additionally, we appreciate the inputs provided by Vivek Srikumar and Ellen Riloff. Vivek Gupta acknowledges support from Bloomberg’s Data Science Ph.D. Fellowship.
\bibliography{anthology,custom}
\bibliographystyle{acl_natbib}

\appendix

\section{Appendix}
\label{appendix}

\subsection{Table Representation}
\label{tab_representation}

We explored two ways to represent table as follows: 

\begin{itemize}
\item \textit{Premise as a paragraph:} 
Instead of using a universal template like "The $key$ of $title$ is $value$", following \cite{neeraja-etal-2021-incorporating}, we use Better Paragraph Representation (BPR) templates based on table categories and keys associated with entity types. In reference to \textit{Breakfast in America} (\cref{tab:example}), the row "\textbf{Released}: \textit{29 March 1979}" is transformed into "The \textit{released} of \textit{Breakfast in America} is \textit{29 March 1979}." using a universal template. "\textit{Breakfast in America} was \textit{released} on \textit{29 March 1979}." using BPR.

\item \textit{Premise as a Linearized Table:} 
In accordance with \cite{DBLP:conf/iclr/ChenWCZWLZW20}, we describe tables as a series of "key : value" tokens. A comma (",") is used to separate multiple values for the same key from one another, while a semi-colon (";") is used to separate rows.

\item \textit{Table Pruning:} For a particular hypothesis, not all of the entries in the premise table are essential. Sometimes, the entire table with the hypothesis as input might be longer than the specified input length of the language model. Inspired by \citet{neeraja-etal-2021-incorporating}, we used alignment methods used in \citet{yadav-etal-2019-alignment,yadav-etal-2020-unsupervised} to remove distracting rows (DRR). By choosing the top 4 rows, we observed that some vital rows are missing for some examples, making the model detect them as neutral, especially in out-of-domain test sets like $\alpha_3$, so we also consider top-8 rows. We use the top 4 and 8 relevant rows from DRR (DRR@4 and DRR@8, respectively) for evaluation.
\end{itemize}


\subsection{Results with Linearized Table}
\label{lin_table}
We experiment with premise as a linearized table and compared our results with \citet{gupta-etal-2020-infotabs}, see~\cref{tab:linear_table}. Our proposed approach was able to outperform the baselines in \citet{gupta-etal-2020-infotabs} by a significant margin.

\begin{table}[!htbp]
\small
\centering
\begin{tabular}{c c c} 
 \toprule
 \textbf{Test Splits} &\textbf{\citet{gupta-etal-2020-infotabs}} &\textbf{Ours} \\\midrule
Dev &\textbf{77.61} &76.7 \\
$\alpha_1$ &75.06 &\textbf{76.1} \\
$\alpha_2$ &69.02 &\textbf{69.6} \\
$\alpha_3$ &64.61 &\textbf{65.3} \\
 \bottomrule
\end{tabular}
\caption{Results on Linearized Table comparing \citet{gupta-etal-2020-infotabs} and our approach (ADAPET)}
\label{tab:linear_table}
\end{table}

\subsection{Reasoning on Entail / Contradict Hypothesis} 
\label{appendix_EvsC}
We also study the classification of Entailed and Contradictory hypotheses when the model is trained and tested on the data without any Neutral hypotheses, see~\cref{tab:EvsC}. We found that DRR@4, DRR@8 representations of premise performs better that BPR because of the less distracting premise.

\begin{table}[!htbp]
\small
\centering
\setlength{\tabcolsep}{4.0pt}
\begin{tabular}{l c c c c} 
\toprule
\textbf{Splits} &\textbf{RoBERTa$_L$+CLS} &\multicolumn{3}{c}{\textbf{ADAPET}} \\
\cmidrule(lr){2-5}
\textbf{} &\textbf{DRR@4} &\textbf{BPR} &\textbf{DRR@4} &\textbf{DRR@8} \\ 
\midrule
Dev &81.5 &83.5 &\textbf{84.3} &82.8 \\
$\alpha_1$ &80.25 &83.8 &\textbf{84.3} &\textbf{84.3} \\
$\alpha_2$ &64.66 &65.9 &66.9 &\textbf{67.7} \\
$\alpha_3$ &76 &75.1 &\textbf{78.5} &77.4 \\
\bottomrule
\end{tabular}
\caption{Results on two label classification (Entailment \& Contradiction).}
\label{tab:EvsC}
\end{table}

\begin{table*}[btp]
\small
\centering
\begin{tabular}{l | l | l}
\toprule
\multicolumn{1}{l}{Perturb} & \multicolumn{1}{c}{Original text} & \multicolumn{1}{c}{Perturbed text} \\ \midrule

\multirow{1}{*}{\textbf{neg+char}} & \multicolumn{1}{l|}{The genres of the album are pop and rock.} & \multicolumn{1}{l}{The ge\textcolor{blue}{j}nres of the al\textcolor{blue}{z}um are \textcolor{blue}{not} p\textcolor{blue}{b}p and rock.} \\
\multirow{1}{*}{\textbf{neg+name}} & \multicolumn{1}{l|}{Peter Henderson's album was recorded in 1979.} & \multicolumn{1}{l}{\textcolor{blue}{John Doe's} album was \textcolor{blue}{not} recorded in 1979.} \\
\multirow{1}{*}{\textbf{num+char}} & \multicolumn{1}{l|}{The album was recorded in 1979.} & \multicolumn{1}{l}{The album was rec\textcolor{blue}{q}orded in the last h\textcolor{blue}{p}lf of \textcolor{blue}{459}.} \\
\multirow{1}{*}{\textbf{num+name}} & \multicolumn{1}{l|}{Peter Henderson's album was recorded in 1979.} & \multicolumn{1}{l}{\textcolor{blue}{John Doe's} album was recorded in \textcolor{blue}{731}.} \\
\multirow{1}{*}{\textbf{num+neg}} & \multicolumn{1}{l|}{The album was released on 29 March 1978.} & \multicolumn{1}{l}{The album was \textcolor{blue}{not} released on 29 March \textcolor{blue}{346}.} \\
\multirow{1}{*}{\textbf{num+para}} & \multicolumn{1}{l|}{The album was recorded in 1979.} & \multicolumn{1}{l}{\textcolor{blue}{In the second part of 1278, the album was recorded.}} \\
\multirow{1}{*}{\textbf{para+name}} & \multicolumn{1}{l|}{Peter Henderson produces only rock albums.} & \multicolumn{1}{l}{\textcolor{blue}{Only rock albums are produced by John Doe.}} \\
\multirow{1}{*}{\textbf{num+para+name}} & \multicolumn{1}{l|}{Peter Henderson's album was recorded in 1979.} & \multicolumn{1}{l}{\textcolor{blue}{The album by John Doe was recorded in 3147.}} \\
 \bottomrule
 \end{tabular}
\caption{More examples of various perturbations used to generate the adversarial test sets based on \cref{tab:example}} 
\label{tab:data_adv_examples_more}
\end{table*}

\subsection{Robustness on Perturbation Set} \label{appendix_robustness}
We evaluate robustness with premise representation. In \cref{addrr8v_results_bpr,addrr8v_results_drr4} we show the performance of the model on the adversarial tests which are trained and tested with BPR, DRR@4 representations of premise. We found the results are similar to the results in \cref{addrr8v_results}.

\subsection{Qualitative Analysis of Perturbation Sets}
\label{appendix_qualit_pertub}

On a randomly sampled subset containing 100 examples from each of the perturbation sets, we task a human evaluator to label them and give a score (out of 5) to the grammar of the hypotheses (see \cref{tab:label_grammer_human}). For most cases, i.e., 11 out of 14, we observe a correct of > 80$\%$ indicating the correction of our adversarial tests. Furthermore, in half of the cases (7/14), the correctness score was above 95$\%$. Grammar analysis shows that most sentences are highly grammatical, with an average score of 4.5/5.0. In the perturbation \textit{"number+paraphrase"} we only observed 77$\%$ of label correctness. This could be due to changing numbers, followed by paraphrasing, which changed some contradiction hypotheses to neutral ones. A similar observation is also observed in \textit{"number+char"} where numbers are modified in character perturbation. We also compare the models' performance on these sampled perturbed sets after human corrections in labels and grammar (see \cref{results_label_correct}). We observed that the performance on these corrected sets is similar to the generated perturbed sets, as in \cref{addrr8v_results_drr4}.

\begin{table}[!htbp]
\small
\centering
\setlength{\tabcolsep}{1.0pt}
\begin{tabular}{l c c} 
\toprule
\textbf{Perturbation} &\textbf{Label Correctness(\%)} &\textbf{Grammar Score} \\\midrule
character &99 &4.46 \\
location &79 &4.5 \\
name &97 &4.5 \\
negation &93 &4.36 \\
number &81 &4.5 \\
paraphrase &89 &4.42 \\
negation+char &88 &4.3 \\
negation+name &96 &4.5 \\
number+char &77 &4.3 \\
number+name &96 &4.5 \\
number+negation &80 &4.44 \\
num+paraphrase &77 &4.48 \\
num+para+name &95 &4.42 \\
paraphrase+name &94 &4.5 \\
\bottomrule
\end{tabular}
\caption{Results on Label Correctness (\% of our generated labels match with human's predictions ) and average Grammar score (out of 5) from human evaluation.}
\label{tab:label_grammer_human}
\end{table}

\begin{table}[!htbp]
\small
\centering
\setlength{\tabcolsep}{4.0pt}
\begin{tabularx}{\columnwidth}{l @{\hspace{1.2\tabcolsep}} l c r r r} 
 \toprule
 \textbf{Type} &\textbf{Input} &\multicolumn{2}{c}{\textbf{RoBERTa$_L$}} &\multicolumn{2}{c}{\textbf{ADAPET}} \\ \midrule
\multicolumn{2}{c}{\bf Top 5 Accuracy} &\textbf{w/o} &\textbf{+CLS} &\textbf{w/o} &\textbf{+CLS}
\\\midrule
\multirow{6}{*}{Factual} &only \textcolor{cadmiumgreen}{E} &50.4 &40.6 &52.4 &46.6 \\
&prem + \textcolor{cadmiumgreen}{E} &72 &45.3 &71.5 &60.7 \\
&only \textcolor{red}{C} &55.2 &37.4 &56 &47.8 \\
&prem + \textcolor{red}{C} &74.6 &39.3 &70.2 &56 \\
&only \textcolor{cadmiumgreen}{E}$\cup$\textcolor{red}{C} &52.7 &39.1 &54.1 &47.2 \\
&prem + \textcolor{cadmiumgreen}{E}$\cup$\textcolor{red}{C} &73.3 &42.5 &70.9 &58.5 \\
\hdashline

\multirow{6}{*}{Relational} &only \textcolor{cadmiumgreen}{E} &64.9 &51.6 &67.3 &57.5 \\
&prem + \textcolor{cadmiumgreen}{E} &70.8 &49.1 &72.2 &66.3 \\
&only \textcolor{red}{C} &64.7 &53.1 &65.8 &57.8 \\
&prem + \textcolor{red}{C} &71.1 &53.3 &72 &62 \\
&only \textcolor{cadmiumgreen}{E}$\cup$\textcolor{red}{C} &64.8 &52.4 &66.5 &57.6 \\
&prem + \textcolor{cadmiumgreen}{E}$\cup$\textcolor{red}{C} &70.9 &51.3 &72.1 &64.1 \\
 \bottomrule
\end{tabularx}
\caption{Top 5 accuracy of Factual \& Relational Knowledge Evaluation on DRR@4.(w/o - no CLS, RoBERTa$_L$+CLS}
\label{tab:top5drr4_factrelknow}
\end{table}

\subsection{Models Knowledge Evaluation}
\label{appendix_know_eval}
We also evaluated the model's knowledge of the top 5 accuracy metric \cref{tab:top5drr4_factrelknow}. The results follow a similar pattern on the top 1 accuracy metric.

\begin{table*}
\small
\centering
\setlength{\tabcolsep}{4.0pt}
\begin{tabular}{lccccccccc}
\toprule
\textbf{Perturb} &\textbf{RoBERTa$_L$} &\multicolumn{4}{c}{\textbf{ADAPET}} &\multicolumn{4}{c}{\textbf{ADAPET+CLS}} \\
\cmidrule(lr){3-6}\cmidrule(lr){7-10}
 &\textbf{+CLS} &\textbf{token} &\textbf{CWWM} &\textbf{+mixNLI} &\textbf{+MNLI} &\textbf{token} &\textbf{CWWM} &\textbf{+mixNLI} &\textbf{+MNLI}\\\midrule
character &62 &\textbf{69} &61 &64 &65 &\textbf{69} &55 &65 &53 \\
location &64 &\textbf{70} &69 &66 &63 &69 &68 &69 &63 \\
name &36 &\textbf{40} &31 &37 &\textbf{40} &35 &41 &35 &36 \\
negation &43 &\textbf{65} &63 &\textbf{65} &59 &57 &55 &55 &58 \\
number &62 &\textbf{69} &69 &68 &69 &68 &66 &59 &54 \\
paraphrase &66 &\textbf{77} &71 &76 &\textbf{77} &70 &68 &74 &71 \\
negation+char &32 &41 &42 &42 &\textbf{44} &43 &30 &4 &39 \\
negation+name &15 &10 &10 &\textbf{18} &13 &16 &9 &12 &12 \\
number+char &5 &50 &54 &55 &\textbf{60} &49 &40 &54 &50 \\
number+name &22 &20 &17 &24 &\textbf{26} &23 &25 &24 &21 \\
number+negation &33 &58 &\textbf{54} &51 &43 &5 &47 &44 &32 \\
num+paraphrase &52 &52 &58 &60 &50 &\textbf{59} &55 &54 &56 \\
num+para+name &\textbf{18} &10 &3 &8 &15 &14 &15 &\textbf{18} &10 \\
paraphrase+name &33 &\textbf{38} &28 &35 &33 &36 &34 &36 &28 \\
\bottomrule
\end{tabular}
\caption{\label{results_label_correct} Adversarial Reasoning results on human corrected perturbation sets with DRR@4 comparing RoBERTa$_L$+CLS, ADAPET, ADAPET+CLS (without pre-training (token, CWWM), with mixNLI, MNLI pre-training). token, CWWM - masking strategies, mixNLI, MNLI pre-training uses RoBERTa style token masking.}
\end{table*}

\subsection{Error Analysis} \label{Error_Analysis}

\begin{figure*}[ht!]
        \centering
            \includegraphics[width=0.45\textwidth]{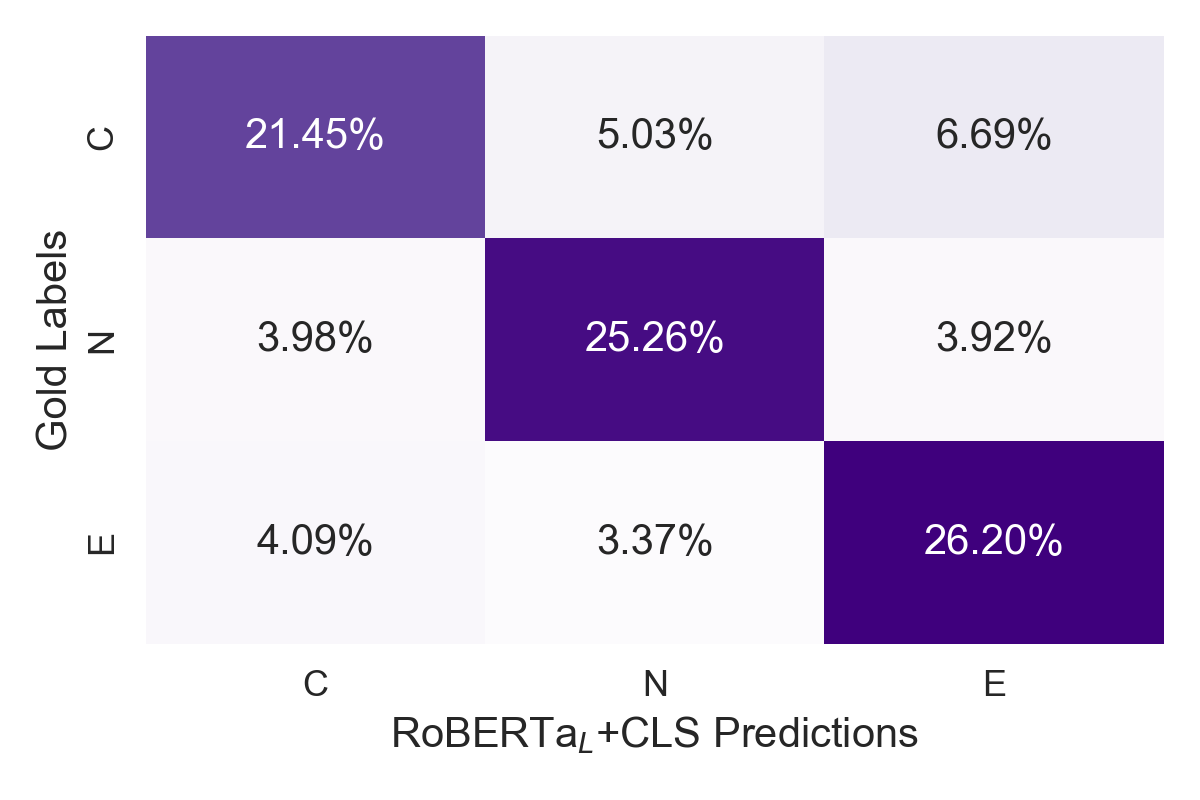}
            \caption{\small Confusion Matrix: Gold Labels vs predictions of RoBERTa$_L$+CLS.}
    \label{fig:confusion_KI}
        \end{figure*}

\begin{figure*}[ht!]
            \includegraphics[width=0.45\textwidth]{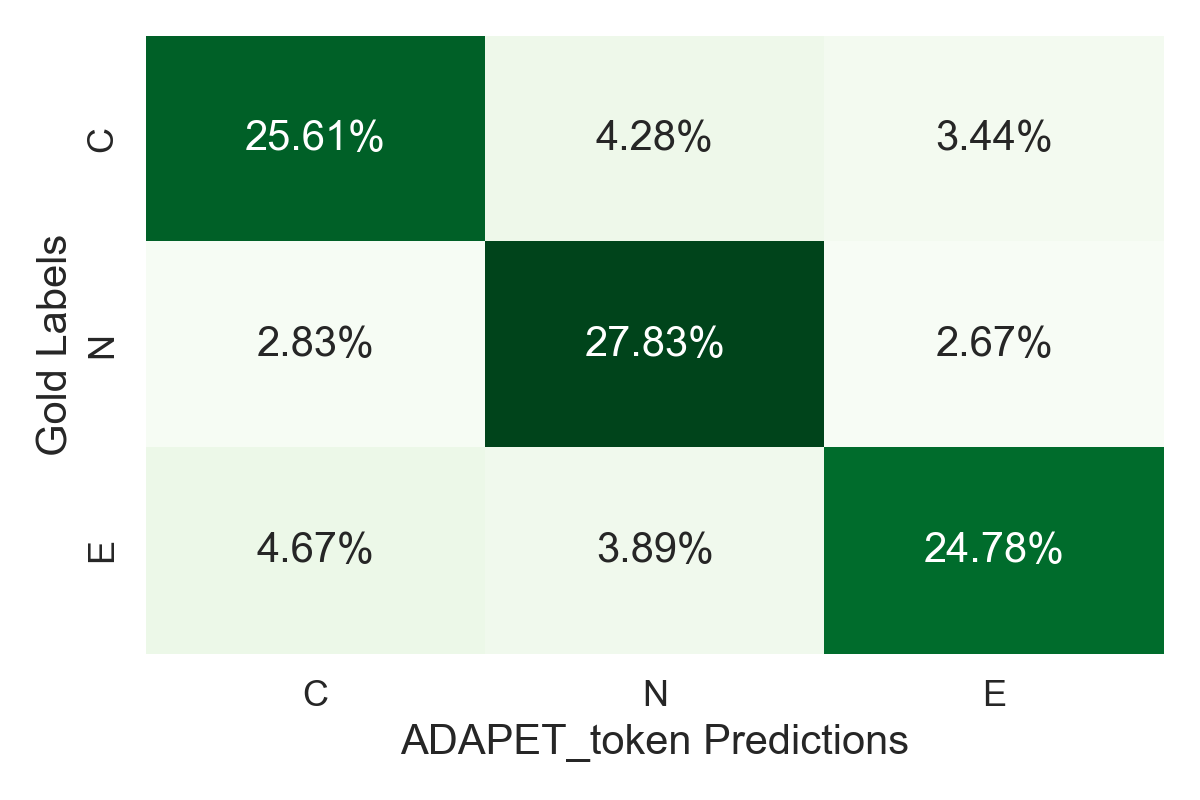}\hfill
            \includegraphics[width=0.45\textwidth]{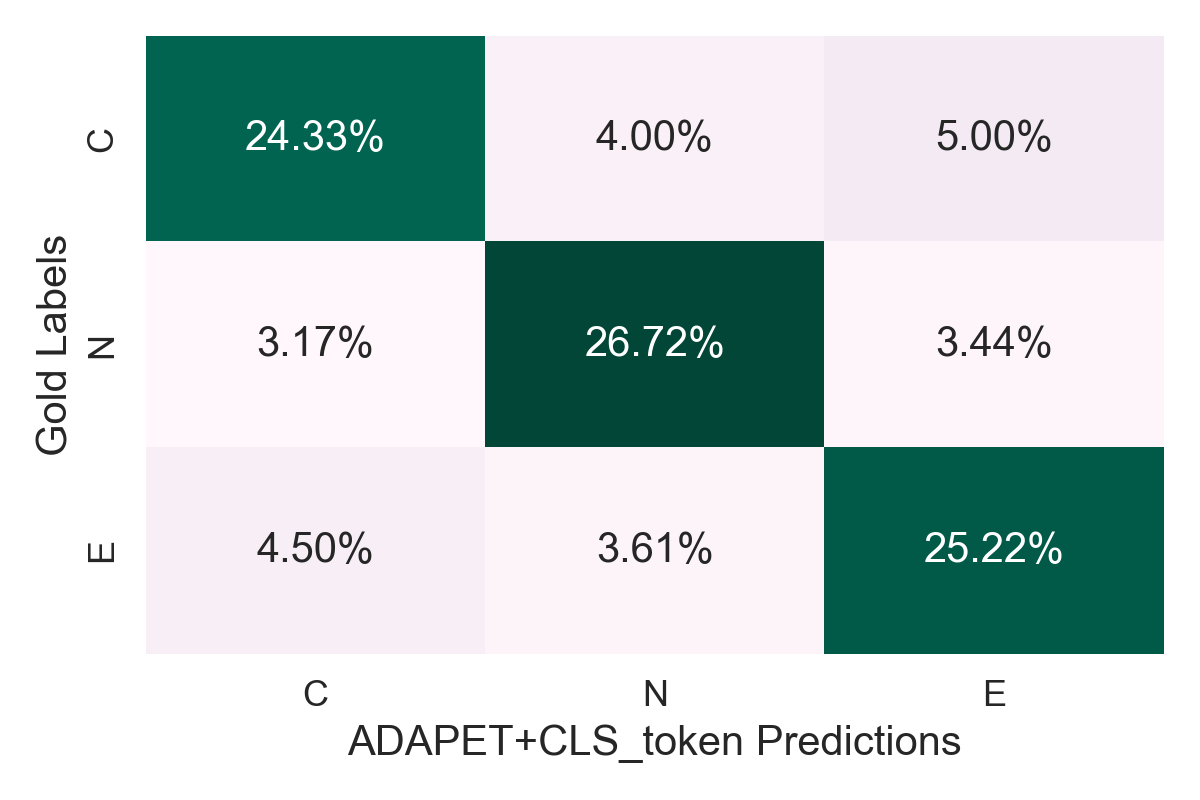}
            \caption{\small Confusion Matrix: Gold Labels vs predictions of ADAPET(token), ADAPET(token)+CLS.}
    \label{fig:confusion_token}
        \end{figure*}

\begin{figure*}[ht!]
            \includegraphics[width=0.45\textwidth]{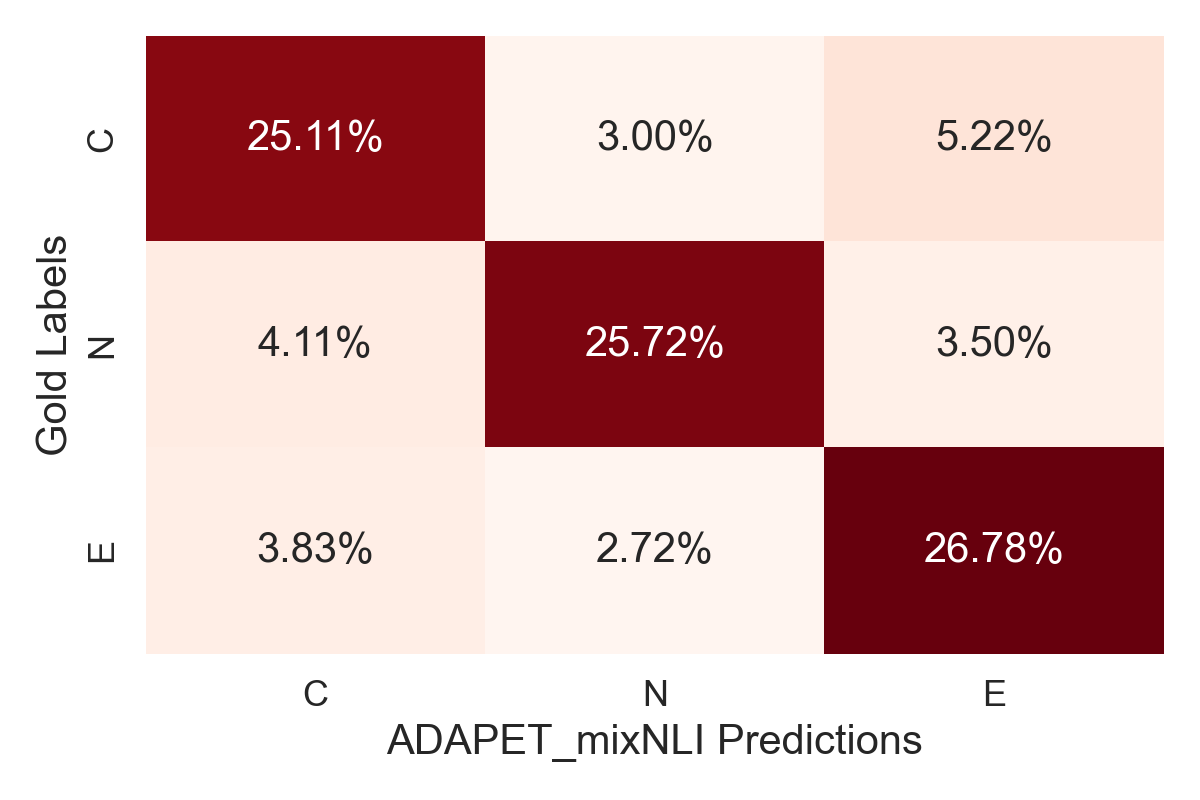}\hfill
            \includegraphics[width=0.45\textwidth]{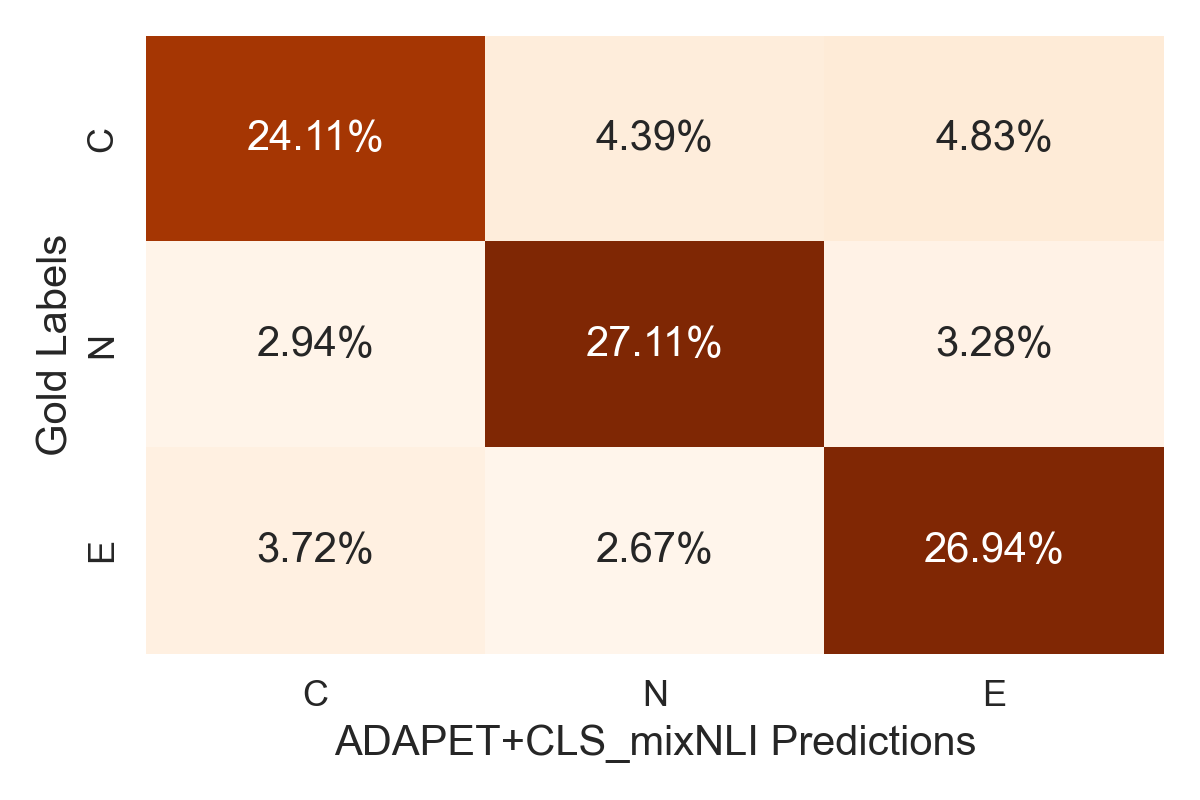}
            \caption{\small Confusion Matrix: Gold Labels vs predictions of ADAPET (pretrained mixNLI), ADAPET (pretrained mixNLI)+CLS.}
    \label{fig:confusion_mixNLI}
        \end{figure*}

In \cref{fig:confusion_token}, when compared to \cref{fig:confusion_KI} there is a substantial improvement in identifying \textcolor{gray}{NEUTRAL} and \textcolor{red}{CONTRADICTION}, but there is also a confusion in  identifying \textcolor{cadmiumgreen}{ENTAILMENT}. Using the NLI-pre-trained model improves the detection of \textcolor{cadmiumgreen}{ENTAILMENT}. A similar observation is also observed with using classifying  layer (+CLS) (see \cref{fig:confusion_token,fig:confusion_mixNLI}).

In \cref{fig:consistency_graph_01}, we see the greatest inconsistency is with \textcolor{gray}{NEUTRAL} being misidentified as \textcolor{cadmiumgreen}{ENTAILMENT} across all models, and this is not that significant with using the classifying layer (+CLS) (see \cref{fig:consistency_graph_02,fig:consistency_graph_04}). Although with the classifying layer, there is increased confusion about \textcolor{red}{CONTRADICTION} being predicted as \textcolor{cadmiumgreen}{ENTAILMENT}.

Table \ref{tab:reason_counts} shows a subset of the validation set labeled based on the different ways the model must think to put the hypothesis in the correct category. On average, all the ADAPET-based models perform similarly, but the human scores are better than the model we utilize. We observe that for certain reasoning types, such as Negation and Simple Look-up, neither humans nor the model arrives at the correct hypothesis, demonstrating the task's difficulty. For Numerical, Lexical, and Entity type reasoning, our model comes very close
to human scores.

In \cref{tab:cat_counts}, we observed that the City category on proposed models performs worse probably as a result of the engagement of more numeric and specific hypotheses compared to the other categories, as well as longer average table size. Our models perform extremely well in identifying \textcolor{cadmiumgreen}{ENTAILMENT} in Food \& Drinks category because of their smaller table size on average and hypothesis requiring no external knowledge to reason as compared to \textcolor{red}{CONTRADICTION}. Our models also struggle in detecting \textcolor{gray}{NEUTRAL} and \textcolor{red}{CONTRADICTION} in Organization category.


\begin{table*}
\small
\centering
\setlength{\tabcolsep}{4.0pt}
\begin{tabular}{lccccccccc}
\toprule
\textbf{Perturb} &\textbf{RoBERTa$_L$} &\multicolumn{4}{c}{\textbf{ADAPET}} &\multicolumn{4}{c}{\textbf{ADAPET+CLS}} \\
\cmidrule(lr){3-6}\cmidrule(lr){7-10}
 &\textbf{+CLS} &\textbf{token} &\textbf{CWWM} &\textbf{+mixNLI} &\textbf{+MNLI} &\textbf{token} &\textbf{CWWM} &\textbf{+mixNLI} &\textbf{+MNLI}\\\midrule
negation+name &11.74 &10.4 &10.2 &\textbf{21.1} &15.6 &17.35 &14.37 &13.89 &12.93 \\
num+para+name &14.06 &10.6 &8.4 &\textbf{20.7} &12 &17.13 &16.88 &14.83 &13.04 \\
number+name &17.26 &12.5 &10.2 &\textbf{20.9} &14.8 &18.42 &18.81 &18.42 &16.88 \\
paraphrase+name &33 &25.8 &20.6 &\textbf{37.6} &31.5 &31.2 &33.41 &32.1 &31.3 \\\hdashline
random &33.33 &33.33 &33.3 &33.33 &33.33 &33.33 &33.33 &33.33 &33.33 \\\hdashline
name &34.6 &26.5 &20.4 &\textbf{36.4} &33.4 &32.41 &34.82 &33.96 &33.2 \\
negation+char &37.71 &38.5 &40.3 &\textbf{47.8} &41.3 &43.56 &40.21 &41.25 &40.49 \\
number+negation &38.36 &30.2 &48.7 &\textbf{54.8} &30.1 &37.69 &47.26 &38.7 &26.06 \\
negation &48.9 &54.2 &57.2 &\textbf{65.4} &55.3 &58.27 &55.27 &58.45 &55.6 \\
number &56.63 &\textbf{62.3} &55.8 &51.9 &56 &55.43 &50.53 &53.52 &56.1 \\
num+paraphrase &56.98 &\textbf{62.3} &57.6 &49.7 &54.5 &55.55 &49.34 &52.26 &55.19 \\
number+char &59.11 &\textbf{66.1} &60.3 &45.1 &55.6 &55.9 &49.32 &52.46 &60.2 \\
character &61.5 &64.1 &62.5 &64.4 &66.1 &64.9 &63.16 &\textbf{66.61} &65.94 \\
location &68.2 &72.4 &\textbf{72.7} &68.1 &70.1 &69.08 &67.69 &66.47 &69.48 \\
paraphrase &68.44 &72.3 &71.8 &\textbf{72.6} &72.3 &72.05 &70.33 &71.7 &\textbf{72.66} \\
\hdashline
dev &76.83 &78.1 &76.4 &\textbf{79.8} &79.1 &78.72 &78.05 &79.22 &78.55 \\
$\alpha_1$ &75.29 &78.1 &76.1 &77.4 &77.4 &77.38 &77.83 &78 &\textbf{78.38} \\
\bottomrule
\end{tabular}
\caption{\label{addrr8v_results_bpr} Adversarial Reasoning results on perturbed sets with BPR comparing RoBERTa$_L$+CLS, ADAPET, ADAPET+CLS (without pre-training (token, CWWM), with mixNLI, MNLI pre-training). token, CWWM - masking strategies, mixNLI, MNLI pre-training uses RoBERTa style token masking. Rows in the tables are sorted in ascending order w.r.t RoBERTa$_L$+CLS performance.}
\end{table*}

\begin{table*}
\small
\centering
\setlength{\tabcolsep}{4.0pt}
\begin{tabular}{lccccccccc}
\toprule
\textbf{Perturb} &\textbf{RoBERTa$_L$} &\multicolumn{4}{c}{\textbf{ADAPET}} &\multicolumn{4}{c}{\textbf{ADAPET+CLS}} \\
\cmidrule(lr){3-6}\cmidrule(lr){7-10}
 &\textbf{+CLS} &\textbf{token} &\textbf{CWWM} &\textbf{+mixNLI} &\textbf{+MNLI} &\textbf{token} &\textbf{CWWM} &\textbf{+mixNLI} &\textbf{+MNLI}\\\midrule
number+name &14.17 &20 &12.9 &14.5 &18.3 &17.78 &17.13 &\textbf{20.8} &16.49 \\
num+para+name &15.08 &16.3 &8.7 &9.5 &15.2 &15.08 &16.88 &\textbf{17.9} &11.25 \\
negation+name &18.66 &17.1 &13.9 &7.8 &11.6 &\textbf{18.48} &13.23 &10.31 &10.55 \\
number+negation &28.63 &36.9 &43.2 &41.5 &23.1 &39.31 &\textbf{45.86} &37.91 &25.78 \\
paraphrase+name &30.9 &32.3 &22.6 &26.7 &27.4 &32.2 &32.36 &\textbf{32.48} &26.55 \\
name &32.4 &32.1 &25.7 &29.8 &30.5 &33.56 &33.6 &\textbf{33.7} &30.01 \\\hdashline
random &33.33 &33.33 &33.33 &33.33 &33.33 &33.33 &33.33 &33.33 &33.33 \\\hdashline
negation+char &40.38 &42.5 &41.1 &39.7 &37.4 &\textbf{45.4} &40.61 &40.49 &38.9 \\
negation &46.46 &\textbf{59.4} &57 &56 &52 &59.03 &56.89 &58.4 &55.7 \\
num+paraphrase &52.56 &57.3 &59.5 &58.4 &\textbf{59.4} &57.7 &51.86 &51.13 &48.9 \\
number+char &53.34 &55.5 &63.2 &61.6 &\textbf{64.8} &55.3 &49.81 &55.85 &54.9 \\
number &54.9 &59.5 &59.1 &56.9 &\textbf{59.8} &55.91 &52.09 &51.97 &51.13 \\
character &56.88 &63.7 &63.7 &\textbf{67.1} &63.3 &65.16 &60.88 &65.16 &65.27 \\
paraphrase &66.3 &72.5 &72.9 &\textbf{73.1} &72.2 &69.88 &68.44 &73.1 &72.22 \\
location &69.65 &\textbf{73} &71.2 &70 &69.9 &69.97 &65.825 &68.59 &68.1 \\
\hdashline
dev &76.39 &76.4 &77.8 &\textbf{78.2} &77.2 &76.27 &78.05 &78.16 &77.5 \\
$\alpha_1$ &75.78 &76.5 &78 &\textbf{79.4} &79.2 &76.44 &77.66 &78.22 &78.11 \\
\bottomrule
\end{tabular}
\caption{\label{addrr8v_results_drr4} Adversarial Reasoning results on perturbed sets with DRR@4 RoBERTa$_L$+CLS, ADAPET, ADAPET+CLS (without pre-training (token, CWWM), with mixNLI, MNLI pre-training). token, CWWM - masking strategies, mixNLI, MNLI pre-training uses RoBERTa style token masking. Rows in the tables are sorted in ascending order w.r.t RoBERTa$_L$+CLS performance.}
\end{table*}

\begin{table*}[!htbp]
\small
\centering
\setlength{\tabcolsep}{0.8pt}
\scriptsize
\begin{tabular}{lcccccccccccccccc}\toprule
\multirow{3}{*}{Reasoning Type} &\multicolumn{5}{c}{\textcolor{cadmiumgreen}{ENTAILMENT}} &\multicolumn{5}{c}{\textcolor{gray}{NEUTRAL}} &\multicolumn{5}{c}{\textcolor{red}{CONTRADICTION}} \\
\cmidrule(lr){2-6} \cmidrule(lr){7-11}\cmidrule(lr){12-16}
&RoBERTa$_L$ &\multicolumn{2}{c}{ADAPET} &\multicolumn{2}{c}{ADAPET+CLS} &RoBERTa$_L$ &\multicolumn{2}{c}{ADAPET} &\multicolumn{2}{c}{ADAPET+CLS} &RoBERTa$_L$ &\multicolumn{2}{c}{ADAPET} &\multicolumn{2}{c}{ADAPET+CLS} \\
\cmidrule(lr){3-4}\cmidrule(lr){5-6} \cmidrule(lr){8-9}\cmidrule(lr){10-11} \cmidrule(lr){13-14}\cmidrule(lr){15-16}
&+CLS &token &+mixNLI &token &+mixNLI &+CLS &token &+mixNLI &token &+mixNLI &+CLS &token &+mixNLI &token &+mixNLI \\\midrule

Numerical (\textcolor{cadmiumgreen}{11}, \textcolor{gray}{3}, \textcolor{red}{7}) &9 &9 &10 &10 &8 &3 &2 &3 &3 &3 &6 &6 &4 &6 &5 \\
Lexical Reasoning (\textcolor{cadmiumgreen}{5}, \textcolor{gray}{3}, \textcolor{red}{4}) &5 &4 &4 &3 &5 &2 &1 &1 &1 &2 &2 &3 &2 &3 &3 \\
Subjective/OOT (\textcolor{cadmiumgreen}{6}, \textcolor{gray}{41}, \textcolor{red}{6}) &3 &3 &3 &3 &3 &37 &36 &36 &37 &35 &4 &4 &1 &3 &5 \\
KCS (\textcolor{cadmiumgreen}{31}, \textcolor{gray}{21}, \textcolor{red}{24}) &25 &21 &26 &20 &25 &20 &20 &18 &19 &18 &21 &22 &18 &21 &21 \\
Temporal (\textcolor{cadmiumgreen}{19}, \textcolor{gray}{11}, \textcolor{red}{25}) &16 &13 &15 &15 &14 &7 &6 &5 &6 &7 &18 &20 &15 &17 &17 \\
Multirow (\textcolor{cadmiumgreen}{20}, \textcolor{gray}{16}, \textcolor{red}{17}) &13 &12 &15 &15 &13 &13 &12 &11 &11 &13 &15 &16 &14 &15 &13 \\
Coref (\textcolor{cadmiumgreen}{8}, \textcolor{gray}{22}, \textcolor{red}{13}) &5 &6 &5 &6 &6 &19 &20 &18 &20 &18 &7 &10 &8 &7 &8 \\
Quantification (\textcolor{cadmiumgreen}{4}, \textcolor{gray}{13}, \textcolor{red}{6})  &2 &2 &2 &2 &2 &11 &11 &12 &12 &12 &2 &3 &3 &3 &3 \\
Named Entity (\textcolor{cadmiumgreen}{2}, \textcolor{gray}{2}, \textcolor{red}{1}) &1 &2 &2 &1 &2 &1 &1 &2 &1 &1 &1 &1 &1 &1 &1 \\
Simple Lookup (\textcolor{cadmiumgreen}{3}, \textcolor{gray}{0}, \textcolor{red}{1}) &2 &3 &3 &2 &3 &0 &0 &0 &0 &0 &0 &0 &0 &0 &0 \\
Negation (\textcolor{cadmiumgreen}{0}, \textcolor{gray}{0}, \textcolor{red}{6}) &0 &0 &0 &0 &0 &0 &0 &0 &0 &0 &4 &6 &5 &5 &4 \\
Entity Type (\textcolor{cadmiumgreen}{6}, \textcolor{gray}{8}, \textcolor{red}{6}) &6 &5 &5 &4 &6 &7 &7 &7 &7 &7 &6 &6 &5 &6 &4 \\
\bottomrule
\end{tabular}
\caption{Reasoning wise number of correct predictions of DRR@4 on subset of dev set, (\textcolor{cadmiumgreen}{a}, \textcolor{gray}{b}, \textcolor{red}{c}) are human prediction count. }\label{tab:reason_counts}
\end{table*}

\begin{table*}[!htbp]
\small
\centering
\setlength{\tabcolsep}{1.5pt}
\scriptsize
\begin{tabular}{lcccccccccccccccc}\toprule
\multirow{3}{*}{Categories} &\multicolumn{5}{c}{\textcolor{cadmiumgreen}{ENTAILMENT}} &\multicolumn{5}{c}{\textcolor{gray}{NEUTRAL}} &\multicolumn{5}{c}{\textcolor{red}{CONTRADICTION}} \\
\cmidrule(lr){2-6} \cmidrule(lr){7-11}\cmidrule(lr){12-16}
&RoBERTa$_L$ &\multicolumn{2}{c}{ADAPET} &\multicolumn{2}{c}{ADAPET+CLS} &RoBERTa$_L$ &\multicolumn{2}{c}{ADAPET} &\multicolumn{2}{c}{ADAPET+CLS} &RoBERTa$_L$ &\multicolumn{2}{c}{ADAPET} &\multicolumn{2}{c}{ADAPET+CLS} \\
\cmidrule(lr){3-4}\cmidrule(lr){5-6} \cmidrule(lr){8-9}\cmidrule(lr){10-11} \cmidrule(lr){13-14}\cmidrule(lr){15-16}
&+CLS &token &+mixNLI &token &+mixNLI &+CLS &token &+mixNLI &token &+mixNLI &+CLS &token &+mixNLI &token &+mixNLI \\\midrule
Album &71 &79 &74 &76 &81 &76 &86 &88 &86 &93 &60 &79 &79 &74 &74 \\
Animal &78 &81 &89 &89 &85 &70 &81 &81 &85 &81 &56 &70 &74 &81 &78 \\
City &59 &63 &63 &57 &69 &67 &80 &65 &71 &75 &53 &61 &63 &65 &55 \\
Country &78 &75 &83 &64 &78 &56 &67 &64 &61 &72 &56 &69 &72 &58 &67 \\
Food\&Drinks &96 &88 &88 &88 &88 &67 &75 &75 &71 &79 &83 &88 &79 &71 &71 \\
Movie &85 &75 &83 &80 &80 &75 &85 &70 &82 &73 &62 &75 &80 &73 &80 \\
Musician &87 &78 &84 &83 &88 &86 &90 &85 &89 &89 &75 &83 &79 &78 &78 \\
Organization &83 &50 &100 &75 &92 &58 &75 &50 &83 &75 &58 &58 &58 &50 &50 \\
Painting &78 &81 &81 &81 &85 &93 &93 &93 &96 &93 &78 &89 &85 &78 &85 \\
Person &74 &73 &78 &74 &78 &81 &85 &80 &78 &81 &67 &79 &76 &77 &74 \\
Others &71 &69 &82 &69 &80 &64 &78 &69 &73 &73 &49 &73 &69 &67 &60 \\
\bottomrule
\end{tabular}
\caption{Category wise accuracy scores of DRR@4 on dev set}\label{tab:cat_counts}
\end{table*}

\begin{minipage}{.94\columnwidth}
\bigskip 
\centering
  {\small 
\begin{tikzpicture}[thick, scale=0.6, {every node/.style}={scale=0.6}]
	\begin{pgfonlayer}{nodelayer}
		\node [style=new style 0] (0) at (0, 8) {C};
		\node [style=new style 0] (1) at (-7, 0) {E};
		\node [style=new style 0] (2) at (7, 0) {N};
		\node [style=none] (3) at (-6.5, 1.5) {};
		\node [style=none] (4) at (-6, 1) {};
		\node [style=none] (5) at (-5.75, -0.5) {};
		\node [style=none] (6) at (-5.75, 0.25) {};
		\node [style=none] (7) at (-1.25, 7.25) {};
		\node [style=none] (8) at (-0.75, 6.75) {};
		\node [style=none] (9) at (0.75, 6.75) {};
		\node [style=none] (10) at (1.25, 7.25) {};
		\node [style=none] (11) at (6, 1) {};
		\node [style=none] (12) at (6.5, 1.5) {};
		\node [style=none] (13) at (5.5, -0.5) {};
		\node [style=none] (14) at (5.5, 0.25) {};
		\node [style=new style 1] (20) at (0, 11.75) {};
\node [style=new style 1] (22) at (7.5, -3.75) {85.17,80.5,86.17};
\node [style=new style 1] (23) at (0, 1) {};
\node [style=new style 1] (24) at (-7.25, -3.75) {};
\node [style=new style 2, rotate=47] (29) at (-4.75, 4.5) {8.83,8.83,6.5};
\node [style=new style 2, rotate=47] (30) at (-3, 3.25) {19.17,8.33,12.67};
\node [style=new style 2] (33) at (0, -1.25) {};
\node [style=new style 2] (48) at (0, 1) {11.33,15.83,10.33};
\node [style=new style 1] (49) at (0, -1.25) {};
\node [style=new style 2] (50) at (0, -1.25) {4.83,0.83,2.67};
\node [style=new style 2] (51) at (0, 11.75) {};
\node [style=new style 1] (52) at (0, 11.75) {68.67,87.33,81.17};
\node [style=new style 1] (53) at (-7.25, -3.75) {79.0,83.0,83.5};
\node [style=new style 2, rotate=-47] (56) at (4.25, 5) {11.5,11.67,11.5};
\node [style=new style 2, rotate=-47] (57) at (2.75, 3.5) {11.5,3.67,5.5};
	\end{pgfonlayer}
	\begin{pgfonlayer}{edgelayer}
		\draw [style=new edge style 0] (8.center) to (4.center); 
		\draw [style=new edge style 0] (14.center) to (6.center); 
		\draw [style=new edge style 3, in=-135, out=-45, loop] (2) to ();
		\draw [style=new edge style 0] (5.center) to (13.center); 
		\draw [style=new edge style 0] (3.center) to (7.center); 
		\draw [style=new edge style 0] (9.center) to (11.center); 
		\draw [style=new edge style 0] (12.center) to (10.center);
		\draw [style=new edge style 3, in=-135, out=-45, loop] (1) to ();
		\draw [style=new edge style 3, in=135, out=45, loop] (0) to ();
	\end{pgfonlayer}
\end{tikzpicture}}
    
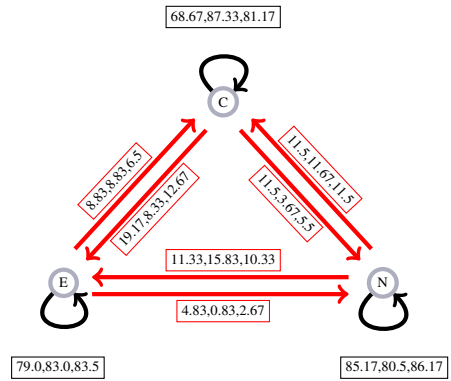
\captionof{figure}{Consistency graph for predictions of ADAPET(token) vs (a) RoBERTa$_L$+CLS (b) ADAPET (CWWM) (c) ADAPET (pretrained mixNLI) in that order respectively.}
    \label{fig:consistency_graph_01}
    \smallskip
\end{minipage}
\begin{minipage}{.94\columnwidth}
\bigskip 
\centering
  {\small 
\begin{tikzpicture}[thick, scale=0.6, {every node/.style}={scale=0.6}]
	\begin{pgfonlayer}{nodelayer}
		\node [style=new style 0] (0) at (0, 8) {C};
		\node [style=new style 0] (1) at (-7, 0) {E};
		\node [style=new style 0] (2) at (7, 0) {N};
		\node [style=none] (3) at (-6.5, 1.5) {};
		\node [style=none] (4) at (-6, 1) {};
		\node [style=none] (5) at (-5.75, -0.5) {};
		\node [style=none] (6) at (-5.75, 0.25) {};
		\node [style=none] (7) at (-1.25, 7.25) {};
		\node [style=none] (8) at (-0.75, 6.75) {};
		\node [style=none] (9) at (0.75, 6.75) {};
		\node [style=none] (10) at (1.25, 7.25) {};
		\node [style=none] (11) at (6, 1) {};
		\node [style=none] (12) at (6.5, 1.5) {};
		\node [style=none] (13) at (5.5, -0.5) {};
		\node [style=none] (14) at (5.5, 0.25) {};
		\node [style=new style 1] (20) at (0, 11.75) {};
\node [style=new style 1] (22) at (7.5, -3.75) {85.17,89.5,89.83};
\node [style=new style 1] (23) at (0, 1) {};
\node [style=new style 1] (24) at (-7.25, -3.75) {};
\node [style=new style 2, rotate=47] (29) at (-4.75, 4.5) {11.67,8.5,7.67};
\node [style=new style 2, rotate=47] (30) at (-3, 3.25) {16.83,9.33,10.33};
\node [style=new style 2] (33) at (0, -1.25) {};
\node [style=new style 2] (48) at (0, 1) {8.17,7.67,7.17};
\node [style=new style 1] (49) at (0, -1.25) {};
\node [style=new style 2] (50) at (0, -1.25) {4.83,5.83,5.67};
\node [style=new style 2] (51) at (0, 11.75) {};
\node [style=new style 1] (52) at (0, 11.75) {67.67,77.33,78.67};
\node [style=new style 1] (53) at (-7.25, -3.75) {84.5,86.67,87.67};
\node [style=new style 2, rotate=-47] (56) at (4.25, 5) {9.67,5.83,6.0};
\node [style=new style 2, rotate=-47] (57) at (2.75, 3.5) {11.5,9.33,7.0};

	\end{pgfonlayer}
	\begin{pgfonlayer}{edgelayer}
		\draw [style=new edge style 0] (8.center) to (4.center); 
		\draw [style=new edge style 0] (14.center) to (6.center); 
		\draw [style=new edge style 3, in=-135, out=-45, loop] (2) to ();
		\draw [style=new edge style 0] (5.center) to (13.center); 
		\draw [style=new edge style 0] (3.center) to (7.center); 
		\draw [style=new edge style 0] (9.center) to (11.center); 
		\draw [style=new edge style 0] (12.center) to (10.center);
		\draw [style=new edge style 3, in=-135, out=-45, loop] (1) to ();
		\draw [style=new edge style 3, in=135, out=45, loop] (0) to ();
	\end{pgfonlayer}
\end{tikzpicture}}
    
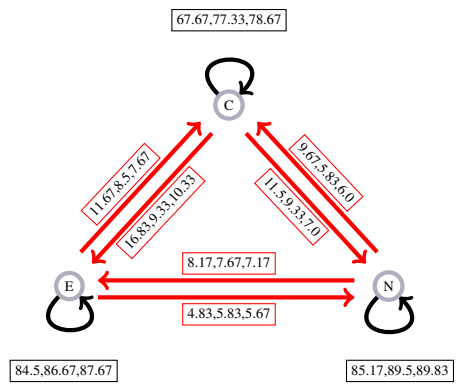
\captionof{figure}{Consistency graph for predictions of ADAPET(token)+CLS vs (a) RoBERTa$_L$+CLS (b) ADAPET (CWWM)+CLS (c) ADAPET (pretrained mixNLI)+CLS in that order respectively.}
    \label{fig:consistency_graph_02}
    \smallskip
\end{minipage}
\begin{minipage}{.94\columnwidth}
\bigskip 
\centering
  {\small 

\begin{tikzpicture}[thick, scale=0.6, {every node/.style}={scale=0.6}]
	\begin{pgfonlayer}{nodelayer}
		\node [style=new style 0] (0) at (0, 8) {C};
		\node [style=new style 0] (1) at (-7, 0) {E};
		\node [style=new style 0] (2) at (7, 0) {N};
		\node [style=none] (3) at (-6.5, 1.5) {};
		\node [style=none] (4) at (-6, 1) {};
		\node [style=none] (5) at (-5.75, -0.5) {};
		\node [style=none] (6) at (-5.75, 0.25) {};
		\node [style=none] (7) at (-1.25, 7.25) {};
		\node [style=none] (8) at (-0.75, 6.75) {};
		\node [style=none] (9) at (0.75, 6.75) {};
		\node [style=none] (10) at (1.25, 7.25) {};
		\node [style=none] (11) at (6, 1) {};
		\node [style=none] (12) at (6.5, 1.5) {};
		\node [style=none] (13) at (5.5, -0.5) {};
		\node [style=none] (14) at (5.5, 0.25) {};
		\node [style=new style 1] (20) at (0, 11.75) {};
\node [style=new style 1] (22) at (7.5, -3.75) {85.17,86.17,86.5};
\node [style=new style 1] (23) at (0, 1) {};
\node [style=new style 1] (24) at (-7.25, -3.75) {};
\node [style=new style 2, rotate=47] (29) at (-4.75, 4.5) {8.83,6.5,5.83};
\node [style=new style 2, rotate=47] (30) at (-3, 3.25) {19.17,12.67,12.33};
\node [style=new style 2] (33) at (0, -1.25) {};
\node [style=new style 2] (48) at (0, 1) {11.33,10.33,11.17};
\node [style=new style 1] (49) at (0, -1.25) {};
\node [style=new style 2] (50) at (0, -1.25) {4.83,2.67,2.83};
\node [style=new style 2] (51) at (0, 11.75) {};
\node [style=new style 1] (52) at (0, 11.75) {68.67,81.17,80.83};
\node [style=new style 1] (53) at (-7.25, -3.75) {79.0,83.5,84.0};
\node [style=new style 2, rotate=-47] (56) at (4.25, 5) {11.5,11.5,10.33};
\node [style=new style 2, rotate=-47] (57) at (2.75, 3.5) {11.5,5.5,6.17};

	\end{pgfonlayer}
	\begin{pgfonlayer}{edgelayer}
		\draw [style=new edge style 0] (8.center) to (4.center); 
		\draw [style=new edge style 0] (14.center) to (6.center); 
		\draw [style=new edge style 3, in=-135, out=-45, loop] (2) to ();
		\draw [style=new edge style 0] (5.center) to (13.center); 
		\draw [style=new edge style 0] (3.center) to (7.center); 
		\draw [style=new edge style 0] (9.center) to (11.center); 
		\draw [style=new edge style 0] (12.center) to (10.center);
		\draw [style=new edge style 3, in=-135, out=-45, loop] (1) to ();
		\draw [style=new edge style 3, in=135, out=45, loop] (0) to ();
	\end{pgfonlayer}
\end{tikzpicture}}
    
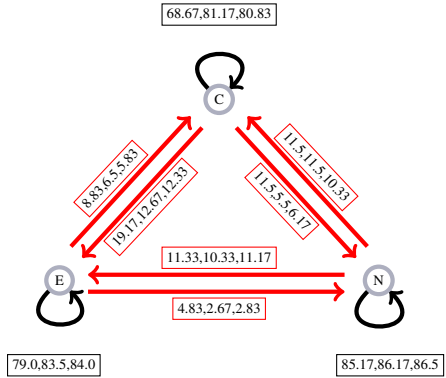
\captionof{figure}{Consistency graph for predictions of ADAPET(token) vs (a) RoBERTa$_L$+CLS (b) ADAPET (pretrained mixNLI) (c) ADAPET (pretrained MNLI) in that order respectively.}
    \label{fig:consistency_graph_03}
    \smallskip
\end{minipage}
\begin{minipage}{.94\columnwidth}
\bigskip 
\centering
  {\small 
\begin{tikzpicture}[thick, scale=0.6, {every node/.style}={scale=0.6}]
	\begin{pgfonlayer}{nodelayer}
		\node [style=new style 0] (0) at (0, 8) {C};
		\node [style=new style 0] (1) at (-7, 0) {E};
		\node [style=new style 0] (2) at (7, 0) {N};
		\node [style=none] (3) at (-6.5, 1.5) {};
		\node [style=none] (4) at (-6, 1) {};
		\node [style=none] (5) at (-5.75, -0.5) {};
		\node [style=none] (6) at (-5.75, 0.25) {};
		\node [style=none] (7) at (-1.25, 7.25) {};
		\node [style=none] (8) at (-0.75, 6.75) {};
		\node [style=none] (9) at (0.75, 6.75) {};
		\node [style=none] (10) at (1.25, 7.25) {};
		\node [style=none] (11) at (6, 1) {};
		\node [style=none] (12) at (6.5, 1.5) {};
		\node [style=none] (13) at (5.5, -0.5) {};
		\node [style=none] (14) at (5.5, 0.25) {};
		\node [style=new style 1] (20) at (0, 11.75) {};
\node [style=new style 1] (22) at (7.5, -3.75) {85.17,89.83,86.17};
\node [style=new style 1] (23) at (0, 1) {};
\node [style=new style 1] (24) at (-7.25, -3.75) {};
\node [style=new style 2, rotate=47] (29) at (-4.75, 4.5) {11.67,7.67,7.67};
\node [style=new style 2, rotate=47] (30) at (-3, 3.25) {16.83,10.33,13.83};
\node [style=new style 2] (33) at (0, -1.25) {};
\node [style=new style 2] (48) at (0, 1) {8.17,7.17,9.17};
\node [style=new style 1] (49) at (0, -1.25) {};
\node [style=new style 2] (50) at (0, -1.25) {4.83,5.67,4.33};
\node [style=new style 2] (51) at (0, 11.75) {};
\node [style=new style 1] (52) at (0, 11.75) {67.67,78.67,75.17};
\node [style=new style 1] (53) at (-7.25, -3.75) {84.5,87.67,89.0};
\node [style=new style 2, rotate=-47] (56) at (4.25, 5) {9.67,6.0,7.67};
\node [style=new style 2, rotate=-47] (57) at (2.75, 3.5) {11.5,7.0,7.0};

	\end{pgfonlayer}
	\begin{pgfonlayer}{edgelayer}
		\draw [style=new edge style 0] (8.center) to (4.center); 
		\draw [style=new edge style 0] (14.center) to (6.center); 
		\draw [style=new edge style 3, in=-135, out=-45, loop] (2) to ();
		\draw [style=new edge style 0] (5.center) to (13.center); 
		\draw [style=new edge style 0] (3.center) to (7.center); 
		\draw [style=new edge style 0] (9.center) to (11.center); 
		\draw [style=new edge style 0] (12.center) to (10.center);
		\draw [style=new edge style 3, in=-135, out=-45, loop] (1) to ();
		\draw [style=new edge style 3, in=135, out=45, loop] (0) to ();
	\end{pgfonlayer}
\end{tikzpicture}}
    \captionof{figure}{Consistency graph for predictions of ADAPET(token)+CLS vs (a) RoBERTa$_L$+CLS (b) ADAPET (pretrained mixNLI)+CLS (c) ADAPET (pretrained MNLI)+CLS in that order respectively.}
    \label{fig:consistency_graph_04}
    \smallskip
\end{minipage}


        \begin{figure*}[ht!]
            \includegraphics[width=0.45\textwidth]{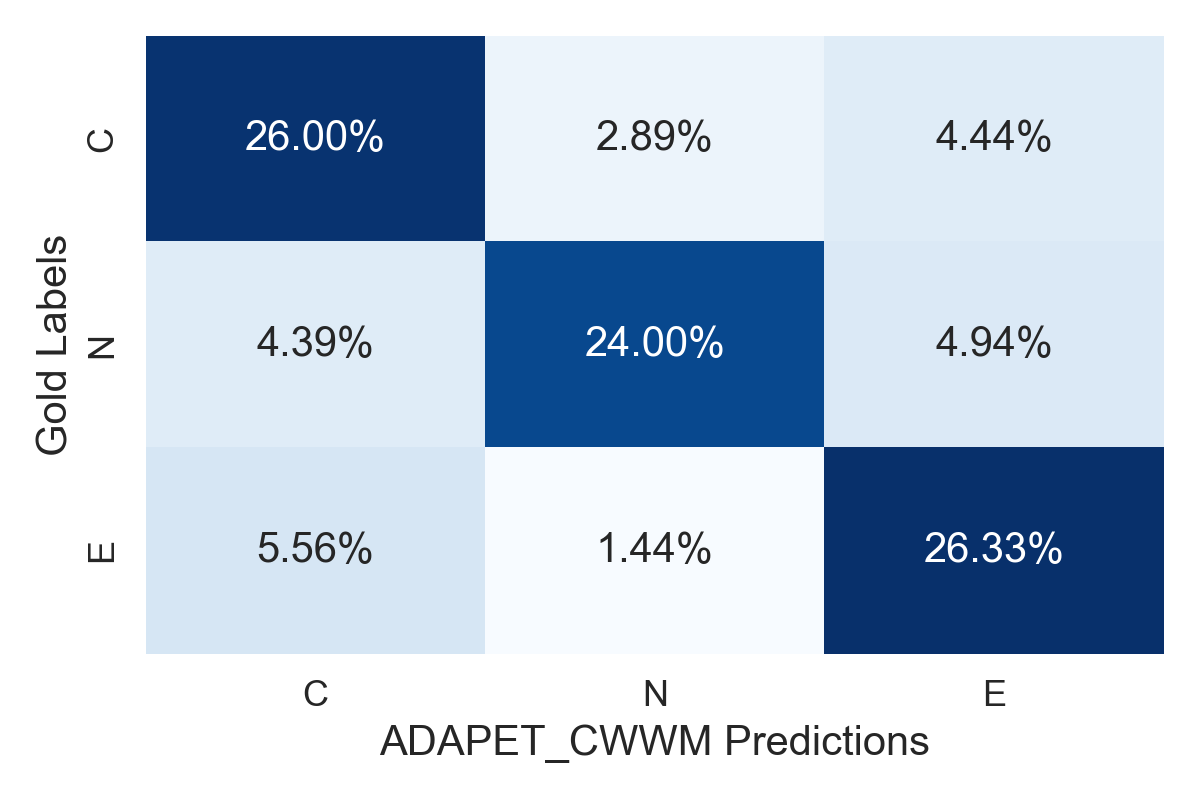}\hfill
            \includegraphics[width=0.45\textwidth]{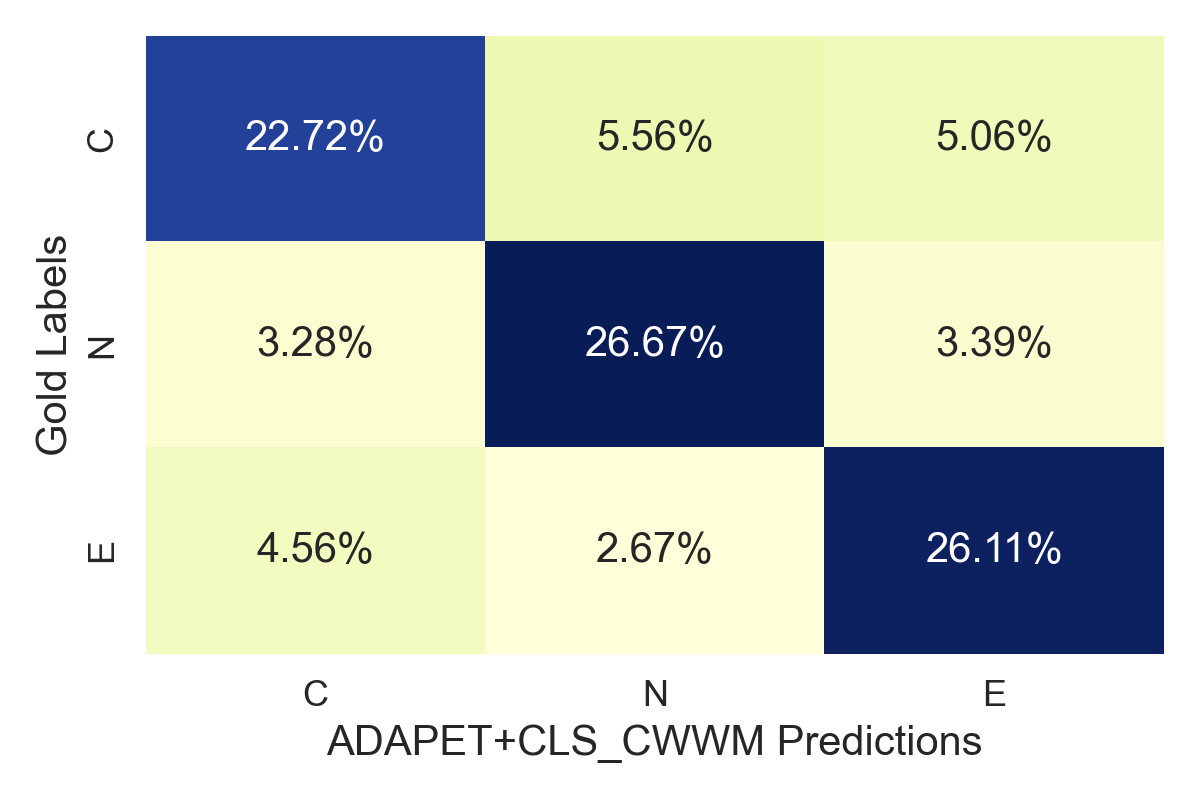}
            \caption{\small Confusion Matrix: Gold Labels vs predictions of ADAPET(CWWM), ADAPET(CWWM)+CLS.}
    \label{fig:confusion_cwwm}
        \end{figure*}
        
\section{Further Discussion}
\label{discussion}
\paragraph{Why table as a paragraph?} A massive data corpus is used to pre-train the large language models. In contrast to semi-structured data, the bulk of pre-training data is unstructured. These models should, of course, perform better on unstructured data and struggle with semi-structured data. Tables in \datasetName~ \cite{gupta-etal-2020-infotabs} are semi-structured in nature. These tables do not explicitly state the relationship between the keys and values; they can also have variable schemas. The album's overall duration is 46:06 minutes, according to the row with key Length and value 46:06. It is difficult to comprehend implicitly that "Length" refers to time length in minutes. Because of the absence of implicit information, a simple table linearization will not be sufficient. \citet{gupta-etal-2020-infotabs,neeraja-etal-2021-incorporating} experimented with various forms of table representations. They found that representing tables as paragraphs gave better results and can leverage the advantage of pre-trained models datasets like MNLI for even better performance.


\paragraph{Why NLI task as cloze-style questions?}
While \citet{gururangan-etal-2018-annotation} showed MLM pre-training with unlabeled target data could further improve the performance on downstream tasks. \citet{DBLP:journals/corr/abs-2110-05301} also showed that using MLM pre-training makes models robust to lexicon-level spurious features. \citet{DBLP:journals/corr/abs-2106-09226} presented a methodology for analysis that connects the pre-training and downstream tasks to an underlying latent variable generative text model. They observed that prompt tuning achieves downstream assurances with less stringent non-degeneracy constraints than head tuning. By reformulating the NLI task as cloze style questions, we can use label conditioned MLM with prompt tuning, which resulted in a better performance on tabular reasoning on \datasetName~.

\end{document}